\documentclass[twoside]{article}

\usepackage[accepted]{aistats2024}

\usepackage{amsfonts,amsmath,amssymb,amsthm,hyperref,caption}
\usepackage{algorithm}
\usepackage{algpseudocode}
\usepackage{tikz}
\usepackage{svg}
\usepackage{amsthm}
\usepackage{booktabs}
\newcommand{\tikzxmark}{%
\tikz[scale=0.23] {
    \draw[line width=0.7,line cap=round] (0,0) to [bend left=6] (1,1);
    \draw[line width=0.7,line cap=round] (0.2,0.95) to [bend right=3] (0.8,0.05);
}}
\newcommand{\tikzcmark}{%
\tikz[scale=0.23] {
    \draw[line width=0.7,line cap=round] (0.25,0) to [bend left=10] (1,1);
    \draw[line width=0.8,line cap=round] (0,0.35) to [bend right=1] (0.23,0);
}}

\newtheorem{thm}{Theorem}
\newtheorem{lemma}{Lemma}

\def\R{\mathbb{ R}}

\def\E{\mathbb{ E}}
\def\calX{\mathcal{ X}}

\def\calD{\mathcal{ D}}
\def\calR{\mathcal{ R}}

\def\calH{\mathcal{ H}}

\def\calU{\mathcal{ U}}
\def\calY{\mathcal{ Y}}

\def\calP{\mathcal{P}}

\def\MMD{\textup{MMD}}

\def\hatpim{\hat{\pi}_m}
\begin{document}

%
\runningtitle{Consistent Optimal Transport with Empirical Conditional Measures}

%
\runningauthor{Piyushi Manupriya, Rachit Keerti Das, Sayantan Biswas, J. SakethaNath}

\twocolumn[

\aistatstitle{Consistent Optimal Transport with Empirical Conditional Measures}

\aistatsauthor{ Piyushi Manupriya\footnotemark[1] \And Rachit Keerti Das\footnotemark[2] \And  Sayantan Biswas\footnotemark[2] \And J. SakethaNath}

\aistatsaddress{
IIT Hyderabad, INDIA \And  Microsoft, INDIA \And  Amazon, INDIA \And IIT Hyderabad, INDIA } ]

\begin{abstract}
Given samples from two joint distributions, we consider the problem of Optimal Transportation (OT) between them when conditioned on a common variable. We focus on the general setting where the conditioned variable may be continuous, and the marginals of this variable in the two joint distributions may not be the same. In such settings, standard OT variants cannot be employed, and novel estimation techniques are necessary. Since the main challenge is that the conditional distributions are not explicitly available, the key idea in our OT formulation is to employ kernelized-least-squares terms computed over the joint samples, which implicitly match the transport plan's marginals with the empirical conditionals. Under mild conditions, we prove that our estimated transport plans, as a function of the conditioned variable, are asymptotically optimal. For finite samples, we show that the deviation in terms of our regularized objective is bounded by $O(1/m^{1/4})$, where $m$ is the number of samples. We also discuss how the conditional transport plan could be modelled using explicit probabilistic models as well as using implicit generative ones. We empirically verify the consistency of our estimator on synthetic datasets, where the optimal plan is analytically known. When employed in applications like prompt learning for few-shot classification and conditional-generation in the context of predicting cell responses to treatment, our methodology improves upon state-of-the-art methods.
\end{abstract}
\footnotetext[1]{Corresponding author (cs18m20p100002@iith.ac.in).}\footnotetext[2]{Work majorly done while at IIT Hyderabad.}

\section{INTRODUCTION}
Optimal Transport (OT) ~\cite{KatoroOT} serves as a powerful tool for comparing distributions. OT has been instrumental in diverse ML applications \cite{CompOT, liu2020semantic,fatras2021jumbot, Cao22,chen2023plot} that involve matching distributions. The need for comparing conditional distributions also frequently arises in machine learning. For instance, in the supervised learning of (probabilistic) discriminative models, one needs to compare the model's label posterior with the label posterior of the training data. Learning implicit conditional-generative models is another such application. Typically, the observed input covariates in these applications are continuous rather than discrete. Consequently, one may only assume access to samples from the input-label joint distribution rather than having multiple samples for a given input. It is well known that estimating conditionals is a significantly more challenging problem than estimating joints (e.g. refer to Section~(2) in ~\cite{LiNeykovBalakrishnan}). Hence, it is not straightforward to apply OT between the relevant conditionals, as the conditionals are implicitly given via samples from the joint distribution.  This issue becomes more pronounced when the distributions of input covariates in the two joints are not the same, e.g. in medical applications~\cite{hahn2019atlantic} where the distributions of treated and untreated patients differ. In such cases, merely performing an OT  between the joint distributions of input and label is not the same as comparing the corresponding conditionals.

In this paper, we address this challenging problem of estimating OT plan between two conditionals, say $s_{Y|X}(\cdot|x)$ and $t_{Y'|X'}(\cdot|x)$, when samples from the joint distributions, $s_{X, Y}, ~t_{X', Y'}$, are given. As motivated above, we do not restrict the conditioned variable to be discrete, nor do we assume that the marginals of the common variable, $s_X$ and $t_{X'}$, are the same. 
As we discuss in our work, the key challenge in estimating OT between conditionals comes in enforcing the marginal constraints involving the conditionals, because the samples provided are not from the conditionals, but from the joints $s_{X, Y}$ and $t_{X', Y'}$. Our formulation employs kernelized-least-squares terms, computed over the joint samples, to address this issue. These regularizer terms implicitly match the transport plan's marginals with the empirical conditionals. Under mild assumptions, we prove that our conditional transport plan is indeed an optimal one, asymptotically. Hence, the corresponding transport cost will match the true Wasserstein between the conditionals. For finite samples, $m$, we show that the deviation in our regularized objective is upper bounded by $O(1/m^{1/4})$.

Few prior works have considered special cases of this problem and have focused on learning conditional optimal transport maps~\cite{Tabak21, Cuturi22}. To the best of our knowledge, our work is the first to formulate OT between conditionals in a general setting that also leads to provably consistent estimators for the optimal transport cost as well as the transport plan as a function of the conditioned variable's value, $x$. Further, instead of directly modelling the transport plan, $\pi_{Y, Y'|X}$, we instead propose modelling it's factors: $\pi_{Y'|Y,X},\ \pi_{Y|X}$. This gives a three-fold advantage: (i) The models for the factors are much simpler than for the joint (ii) when dealing with discriminative/conditional-generative models we can directly choose $\pi_{Y|X}(\cdot|x)$ as the discriminative model being learnt. (ii) When implicit generative models are used for the factors, $\pi_{Y'|Y,X}(\cdot|y,x)$ can be readily be used for inference in applications like cell population dynamics (e.g., section~\ref{sec:simbio}). 

We empirically show the utility of our approach in the conditional generative task for modelling cell population dynamics, where we consistently outperform the baselines. Furthermore, we pose the task of learning prompts for few-shot classification as a conditional optimal transport problem. We argue that this is advantageous than posing it as a classical optimal transport problem, which is the approach existing works employ. We test this novel approach on the benchmark EuroSAT~\cite{helber2019eurosat} dataset and show improvements over~\cite{chen2023plot}, a state-of-the-art prompt learning method.

In Table~\ref{table:relw}, we highlight some of the key features of COT, comparing it with the related works. Our main contributions are summarized below.

\paragraph{Contributions}
\begin{itemize}
\item We propose novel estimators for optimal transport between conditionals in a general setting where the conditioned variable may be continuous, and its marginals in the two joint distributions may differ.
\item We prove the consistency of the proposed estimators. To the best of our knowledge, we are the first to present a consistent estimator for conditional optimal transport in the general setting.
\item While recent approaches model the optimal transport map~\cite{Tabak21},~\cite{Cuturi22}, we model the transport plan, which enables more general inferences.
\item We empirically verify the correctness of the proposed estimator on synthetic datasets. We further evaluate the proposed approach on downstream applications of conditional generation for modelling cell population dynamics and prompt learning for few-shot classification, showing its utility over some of the state-of-the-art baselines.
\end{itemize}
\section{PRELIMINARIES}\label{sec:pre}
Let $\calX,\calY$ be two sets (domains) that form compact Hausdorff spaces. Let $\calP(\calX)$ be the set of all probability measures over $\calX$. 
\newline
\textbf{Optimal Transport (OT)} Given a cost function, $c:\calY\times\calY\mapsto\R$, OT compares two measures $s,t\in\calP(\calY)$ by finding a plan to transport mass from one to the other, that incurs the least expected cost. More formally, Kantorovich's OT formulation~\cite{KatoroOT} is given by:
\begin{equation}\label{eqn:ot}
W_c(s,t)\equiv\min_{\pi\in\calP(\calY\times\calY)}\int c\ \textup{d}\pi, \textup{ s.t.}\ \ \pi_1=s,\ \pi_2=t,
\end{equation}
where $\pi_1,\pi_2$ are the marginals of $\pi$. A valid cost metric over $\calY\times \calY$ defines the $1$-Wasserstein metric, $W_c(s,t)$, over distributions $s,t\in\calP(\calY)$. The cost metric is referred to as the ground metric.
\paragraph{Maximum Mean Discrepancy (MMD)}  Given a characteristic kernel function \cite{characteristicK}, $k:\calY \times \calY \mapsto \R$, MMD defines a metric over probability measures given by: $\MMD^2(s, t) \equiv \mathbb{E}_{X\sim s,X^{'}\sim s}[k(X, X^{'})] + \mathbb{E}_{Y\sim t,Y^{'}\sim t}[k(Y, Y^{'})] -2\mathbb{E}_{X\sim s, Y\sim t}[k(X, Y)]$. With $\calH_k$ as the RKHS associated with the characteristic kernel $k$, the dual norm definition of MMD is given by $\MMD(s, t)=\max_{f\in \calH_k; \|f\|\leq 1}\E_s[f(X)]-\E_t[f(Y)]$.

\begin{table*}
  \caption{Summary of related works and the proposed COT method.}
  \label{table:relw}
  \begin{center}
  \scriptsize{\begin{tabular}{lcccc}
    \toprule
     & \cite{Tabak21} & \cite{Bures} & \cite{Cuturi22} & COT\\
    \midrule
    Consistent estimator & N/A & N/A & N/A & \tikzcmark\\
    Models OT plan with flexibility of implicit modelling & \tikzxmark & \tikzxmark & \tikzxmark & \tikzcmark\\
    Flexibility with the ground cost & \tikzcmark & \tikzxmark & \tikzxmark & \tikzcmark\\
    Allows single sample per conditioned variable & \tikzcmark & \tikzcmark  & \tikzxmark & \tikzcmark \\
    \bottomrule
  \end{tabular}}
  \end{center}
\end{table*}
\section{RELATED WORK}
Few prior works have attempted to solve the conditional OT problem in some special cases, which we discuss below. ~\cite{Frogner15} presents an estimator for the case when the marginals, $s_X$ and $t_{X'}$, are the same and $y$ takes discrete values. Their estimator does not generalize to the case where $y$ is continuous. Further, they solve individual OT problems at each $x$ rather than modelling the transport map/plan as a function of $x$. \cite{Bures} characterizes the conditional distribution discrepancy using the Conditional Kernel Bures (CKB). With the assumption that the kernel embeddings for the source and target are jointly Gaussian, CKB defines a metric between conditionals. \cite{Bures} does not discuss any (sufficient) conditions for this assumption to hold. Moreover, CKB only estimates the discrepancy between the two conditionals, and it is unclear how to retrieve an optimal transport plan/map with CKB, limiting its applications. \cite{Cuturi22} considers special applications where multiple samples from $s_{Y|X}(\cdot|x), ~t_{Y'|X'}(\cdot|x)$ are available at each $x$. They learn a transport map as a function of $x$ by solving standard OT problems between $s_{Y|X}(\cdot|x), ~t_{Y'|X'}(\cdot|x)$ individually for each sample $x$. Also, their approach additionally assumes the ground cost is squared Euclidean. In contrast, we neither assume access to multiple samples from $s_{Y|X}(\cdot|x), ~t_{Y'|X'}(\cdot|x)$ at each $x$ nor make restrictive assumptions on the ground cost. Further, we estimate the transport plan rather than the transport map. The work closest to ours is~\cite{Tabak21}. However, there are critical differences between the two approaches, which we highlight below. \cite{Tabak21} formulates a min-max adversarial formulation with a KL divergence-based regularization to learn a transport map. Such adversarial formulations are often unstable, and~\cite{Tabak21} does not present any convergence results. Their empirical evaluation is also limited to small-scale qualitative experiments. Moreover, unlike the estimation bounds we prove,~\cite{Tabak21} does not discuss any learning theory bounds or consistency results. It is expected that such bounds would be cursed with dimensions~\cite{sduot, sliced-uot}. Additionally, the proposed formulation allows us to learn transport plans using implicit models ($\S$~\ref{imp}). Such an approach may not be possible with KL-regularized formulation in~\cite{Tabak21} due to non-overlapping support of the distributions. Owing to these differences, our proposed method is more widely applicable.

\section{PROBLEM FORMULATION}\label{sec:main}
This section formally defines the Conditional Optimal Transport (COT) problem and presents a consistent estimator for it in the general setting. We begin by recalling the definition of OT between two given measures $s_{Y|X}(\cdot|x)$ and $t_{Y'|X'}(\cdot|x)$ for a given $x$. $W_c\left(s_{Y|X}(\cdot|x), t_{Y'|X'}(\cdot|x)\right)$ is defined as follows.
\begin{align}\label{eqn:indcot}
&\min_{\pi_{Y, Y'|X}(\cdot, \cdot|x)\in\calP(\calY\times\calY)}\int_{\calY\times\calY} c\ \textup{d}\pi_{Y, Y'}(\cdot, \cdot|x), \\
&\textup{ s.t.}\ \ \pi_{Y|X}(\cdot|x)=s_{Y|X}(\cdot|x), \ \pi_{Y'|X}(\cdot|x)=t_{Y'|X'}(\cdot|x),\nonumber
\end{align}
where $\pi_{Y|X}(\cdot|x)$ and $\pi_{Y'|X}(\cdot|x)$ denotes the marginals of $\pi_{Y, Y'|X}(\cdot, \cdot|x)$.
If the cost is a valid metric, then $W_c\left(s_{Y|X}(\cdot|x), ~t_{Y'|X'}(\cdot|x)\right)$ is nothing but the Wasserstein distance between $s_{Y|X}(\cdot|x)$ and $t_{Y'|X'}(\cdot|x)$. While $W_c\left(s_{Y|X}(\cdot|x), ~t_{Y'|X'}(\cdot|x)\right)$ helps comparing/transporting measures given a specific $x\in\calX$, in typical learning applications, one needs a comparison in an expected sense rather than at a specific $x\in\calX$. Accordingly, we consider $\E_{X^{''}\sim a}\left[W_c\left(s_{Y|X}(\cdot|X^{''}), ~t_{Y'|X'}(\cdot|X^{''})\right)\right]$, where $a$ is a given auxiliary measure:
\begin{align}\label{eqn:expcot}
\int_{\calX}&\min_{\begin{array}{c}\scriptstyle
     \pi_{Y, Y'|X}(\cdot, \cdot|x)\in\calP(\calY\times\calY) \\ \scriptstyle
      \forall x\in \calX
\end{array}}\int_{\calY\times\calY} c\ \textup{d}\pi_{Y, Y'|X}(\cdot, \cdot|x) \ \textup{d}a(x), \nonumber\\
&\qquad\quad\qquad\textup{ s.t.}~ \pi_{Y|X}(\cdot|x)=s_{Y|X}(\cdot|x), \nonumber\\ 
&\qquad\quad\qquad\qquad\pi_{Y'|X}(\cdot|x)=t_{Y'|X'}(\cdot|x)\ \forall x\in \calX\nonumber \\
\equiv &\min_{\pi_{Y, Y'|X}:\calX\mapsto\calP(\calY\times\calY)}\int_{\calX}\int_{\calY\times\calY} c\ \textup{d}\pi_{Y, Y'|X}(\cdot, \cdot|x)\ \textup{d}a(x), \nonumber\\
&\qquad\quad\qquad\textup{ s.t.}~ \pi_{Y|X}(\cdot|x)=s_{Y|X}(\cdot|x), \nonumber\\ 
&\qquad\quad\qquad\qquad\pi_{Y'|X}(\cdot|x)=t_{Y'|X'}(\cdot|x)\ \forall x\in \calX.
\end{align}
In the special case where the auxiliary measure, $a$, is degenerate, (\ref{eqn:expcot}) gives back (\ref{eqn:indcot}). 
Henceforth, we analyze the proposed COT formulation defined in (\ref{eqn:expcot}).

Now, in typical machine learning applications, the conditionals are not explicitly given, and only samples from the joints are available. Estimation of COT from samples seems challenging because the problem of estimating conditional densities itself has been acknowledged to be a significantly difficult one with known impossibility results (e.g., refer to Section~2 in 
\cite{LiNeykovBalakrishnan}). Hence, some regularity assumptions are necessary for consistent estimation. Further, even after making appropriate assumptions, the typical estimation errors are cursed with dimensions (e.g., Theorem~2.1 in 
~\cite{Graham2020MinimaxRA}).

On the other hand, estimation of RKHS embeddings of conditional measures can be performed at rates $O(1/m^{1/4})$, where $m$ is the number of samples~\cite{Song2009HilbertSE} \cite{gretton}. This motivates us to enforce the constraints in COT (\ref{eqn:expcot}) by penalizing the distance between their RKHS embeddings. More specifically, we exploit the equivalence: $\pi_{Y|X}(\cdot|x)=s_{Y|X}(\cdot|x)\ \forall x\in \calX\iff\int_\calX\MMD^2\left(\pi_{Y|X}(\cdot|x), ~s_{Y|X}(\cdot|x)\right)\textup{d}s_X(x)=0$. This is true because $\MMD$ is a valid metric and we assume $s_X(x)>0, t_{X'}(x)>0\ \forall\ x\in\calX$. Using this, COT (\ref{eqn:expcot}) can be relaxed as:
\begin{align}\label{eqn:regcot1}
&\min_{\pi_{Y, Y'|X}:\calX\mapsto\calP(\calY\times\calY)}\int_{\calX}\int_{\calY\times\calY} c\ \textup{d}\pi_{Y, Y'|X}(\cdot, \cdot|x)\ \textup{d}a(x) \nonumber\\&\quad + \lambda_1\int_\calX\MMD^2\left(\pi_{Y|X}(\cdot|x), s_{Y|X}(\cdot|x)\right)\textup{d}s_X(x)\nonumber\\
&\quad+\lambda_2\int_\calX\MMD^2\left(\pi_{Y'|X}(\cdot|x), t_{Y'|X'}(\cdot|x)\right)\textup{d}t_{X'}(x),
\end{align}
where $\lambda_1,\lambda_2>0$ are regularization hyperparameters. Note that (\ref{eqn:regcot1}) is exactly the same as (\ref{eqn:expcot}) if $\lambda_1,\lambda_2\rightarrow\infty$. 

Now, we use a standard result, $\E\left[\|G-h(X)\|^2\right]= \E\left[\|G-\E[G|X]\|^2\right] + \E\left[\|\E[G|X] - h(X)\|^2\right]$ with $G$ taken as the kernel mean embedding of $\delta_Y$ and $h(X)$ taken as the kernel mean embedding of $\pi_{Y|X}(\cdot|X)$ \cite{MAL-060}. This gives us $\int_{\calX\times\calY}\MMD^2\left(\pi_{Y|X}(\cdot|x), \delta_y\right)\textup{d}s_{X, Y}(x,y)=\int_\calX\MMD^2\left(\pi_{Y|X}(\cdot|x), ~s_{Y|X}(\cdot|x)\right)\textup{d}s_X(x)+v(s)$, where $v(s)\geq 0$.
Here, $\phi$ is the feature map corresponding to the kernel defining the MMD. This leads to the following formulation:
\begin{align}\label{eqn:regcot2}
&\min_{\pi_{Y, Y'|X}:\calX\mapsto\calP(\calY\times\calY)}\int_{\calX}\int_{\calY\times\calY} c\ \textup{d}\pi_{Y, Y'|X}(\cdot, \cdot|x)\ \textup{d}a(x) \nonumber\\
&\quad+\lambda_1\int_{\calX\times\calY}\MMD^2\left(\pi_{Y|X}(\cdot|x), \delta_y\right)\textup{d}s_{X, Y}(x,y)\nonumber\\
&\quad+\lambda_2\int_{\calX\times\calY}\MMD^2\left(\pi_{Y'|X}(\cdot|x), \delta_y\right)\textup{d}t_{X', Y'}(x,y).
\end{align}
Since $v(s),v(t)$ are independent of $\pi$, the solutions of (\ref{eqn:regcot2}) are exactly the same as those of COT (\ref{eqn:expcot}) as $\lambda_1,\lambda_2\rightarrow\infty$. The advantage of this reformulation is that it can be efficiently estimated using samples from the joints, as we detail below.

\subsection{Sample-Based Estimation}
In our set-up, in order to solve (\ref{eqn:regcot2}) and perform estimation, we are only provided with samples $\calD_m^s = \{(x_1,y_1),\ldots,(x_m,y_m)\}$ and $\calD_m^t=\{(x'_1,y'_1),\ldots,(x'_m,y'_m)\}$ from $s_{X, Y}$ and $t_{X', Y'}$, respectively. Hence, we employ a sample-based estimator for the regularizer terms: $\int_{\calX\times\calY}\MMD^2\left(\pi_{Y|X}(\cdot|x), \delta_y\right)\textup{d}s_{X, Y}(x, y) \approx\frac{1}{m}\sum_{i=1}^m\MMD^2\left(\pi_{Y|X}(\cdot|x_i), \delta_{y_i}\right)$. The following lemma shows that this regularizer estimator is statistically consistent.
\begin{lemma}\label{lemma}
Assuming $k$ is a normalized characteristic kernel, with probability at least $1-\delta$, we have
\begin{align*}
&\scriptstyle{\Big|\int_{\calX\times\calY}\MMD^2\left(\pi_{Y|X}(\cdot|x), \delta_y\right)~ \textup{d}s_{X, Y}(x, y)}\\
&\scriptstyle{-\frac{1}{m}\sum\limits_{i=1}^m\MMD^2\left(\pi_{Y|X}(\cdot|x_i), \delta_{y_i}\right) \Big|
\le 2\sqrt{\frac{2}{m}\log\left(\frac{2}{\delta}\right)}}.
\end{align*}
\end{lemma}

Using this result for the regularization terms, (\ref{eqn:regcot2}) can in-turn be estimated as:
\begin{align}\label{eqn:empcot2}
\scriptstyle{\min\limits_{\pi_{Y, Y'|X}:\calX\mapsto\calP(\calY\times\calY)}}&\scriptstyle{\int_{\calX}\int_{\calY\times\calY} c\ \textup{d}\pi_{Y, Y'|X}(\cdot, \cdot|x)\textup{d}a(x) } \nonumber \\
&\scriptstyle{+\lambda_1\frac{1}{m}\sum_{i=1}^m\MMD^2\left(\pi_{Y|X}(\cdot|x_i), \delta_{y_i}\right)}\nonumber\\
&\scriptstyle{+\lambda_2\frac{1}{m}\sum_{i=1}^m\MMD^2\left(\pi_{Y'|X}(\cdot|x'_i), \delta_{y'_i}\right)}.
\end{align}
We choose not to estimate the first term with empirical average as $a$ is a known distribution. In the following theorem, we prove the consistency of our COT estimator.
\begin{thm}\label{thm1}
Let $\Pi$ be a given model for the conditional transport plans, $\pi_{Y,Y'|X}:\calX\mapsto\calP(\calY\times\calY)$. Assume $\lambda_1=\lambda_2=\lambda$. Let $\hat{\pi}_m, ~\pi^*$ denote optimal solutions over the restricted model $\Pi$ corresponding to (\ref{eqn:empcot2}),(\ref{eqn:regcot2}) respectively. Let $\calU[\hat{\pi}_m],\calU[\pi]$ denote the objectives as a function of $\pi\in\Pi$ in  (\ref{eqn:empcot2}),(\ref{eqn:regcot2}) respectively. Then, we prove the following:
\begin{enumerate}
\item With probability at least $1-\delta$, $\calU[\hat{\pi}_m]-\calU[\pi^*]\le2\lambda_1\calR_{m}(\Pi) +2\lambda_2\calR'_{m}(\Pi)+ 6(\lambda_1+\lambda_2)\sqrt{\frac{2}{m}\log{\frac{3}{\delta}}}$, where the Rademacher based complexity term, $\calR_{m}(\Pi)$, is defined as: $\frac{1}{m}\E\left[\max\limits_{\pi\in\Pi}\sum_{i=1}^m\epsilon_i\MMD^2\left(\pi_{Y|X}(\cdot|X_i), \delta_{Y_i}\right)\right]$; $(X_i,Y_i)$ are IID samples from $s_{X, Y}$ and $\epsilon_i$ denotes the Rademacher random variable. $\calR'_{m}(\Pi)$, is analogously defined as: $\frac{1}{m}\E\left[\max\limits_{\pi\in\Pi}\sum_{i=1}^m\epsilon_i\MMD^2\left(\pi_{Y'|X}(\cdot|X'_i), \delta_{Y'_i}\right)\right]$, where $(X'_i,Y'_i)$ are IID samples from $t_{X', Y'}$ and $\epsilon_i$ denotes the Rademacher random variable. Recall that $\pi_{Y|X}(\cdot|x)$ and $\pi_{Y'|X}(\cdot|x)$ denote the marginals of $\pi_{Y,Y'|X}(\cdot, \cdot|x)$.
\item In the special case $\Pi$ is a neural network based conditional generative model, the kernel employed is universal, normalized, and non-expansive~\cite{pmlr-v168-waarde22a}, and $\lambda=O(m^{1/4})$, with high probability we have that $\calU[\hat{\pi}_m]-\calU[\pi^*]\le O(1/m^{1/4})$. More importantly, when $m\rightarrow\infty$, $\hat{\pi}_m$ is an optimal solution to the original COT problem (\ref{eqn:expcot}) whenever $\Pi$ is rich enough such that $\exists\pi^*\in\Pi\ni\pi^*_{Y|X}(\cdot|x)=s_{Y|X}(\cdot|x)$ and $\pi^*_{Y'|X}(\cdot|x)=t_{Y'|X'}(\cdot|x) \ \forall x\in \calX$.
\end{enumerate}
\end{thm}
The proof is presented in Supplementary~$\S~(\textup{S}1.2)$. The conditions for consistency are indeed mild because (i) neural conditional generators are known to be universal (Lemma 2.1~in~\cite{Liu2021WassersteinGL},
 ~\cite{pmlr-v125-kidger20a})
 (ii) the popularly used Gaussian kernel is indeed universal, normalized, and non-expansive (for a large range of hyperparameters). The proof for the first part of the theorem is an adaptation of classical uniform convergence based arguments; however, further bounding the complexity terms in the case of neural conditional generative models is novel and we derive this using vector contraction inequalities along with various properties of the kernel.

\subsection{Modelling the Transport Plan}\label{modelling}
We now provide details of modelling the transport plan function, i.e., choices for $\Pi$, from a pragmatic perspective. Firstly, we model the transport plan $\pi_{Y, Y'|X}(y, y'|x)$ by modelling its factors: $\pi_{Y'|Y, X}(y'|y, x)$ and $\pi_{Y|X}(y|x)$. Since the factors can be modelled using simpler models, this brings us computational benefits, among other advantages that we discuss. Secondly, employing COT with such a factorization enables us to directly choose $\pi_{Y|X}(\cdot|x)$ as the label posterior of the model to be learnt in discriminative modelling applications. Moreover, the other factor $\pi_{Y'|Y, X}(\cdot|y, x)$ can be readily used for inference (see {$\S$~\ref{sec:simbary},~$\S$~\ref{sec:simbio}}).

\subsubsection{Transport Plan with Explicit Models}\label{expl}
Here, we discuss our modelling choice with explicit probabilistic models when $\calY=\left\{l_1,\ldots,l_n\right\}$ is a finite set. Accordingly, we model the factors $\pi_{Y'|Y, X}(y'|y, x),~\pi_{Y|X}(y|x)$ with fixed-architecture neural networks, parameterized by $\psi$ and $\phi$ respectively, with the output layer as softmax over $|\calY|$ labels. The COT estimator \ref{eqn:empcot2} in this case simplifies as: 
\begin{align}\label{eqn:expexpcot}\nonumber
&\scriptstyle{\min\limits_{\psi,\theta}\int_{\calX}\sum_{i=1,j=1}^{i=n,j=n} c(l_i,l_j)\pi_\psi(l_i|l_j,x)\pi_\theta(l_j|x)\textup{d}a}(x) \\&\qquad\scriptstyle{+\lambda_1\frac{1}{m}\sum_{i=1}^m\MMD^2\left(\sum_{j=1}^n\pi_\psi(\cdot|l_j,~x_i)\pi_\theta(l_j|x_i), ~\delta_{y_i}\right)}\nonumber\\&\qquad\scriptstyle{+\lambda_2\frac{1}{m}\sum_{i=1}^m\MMD^2\left(\pi_\theta(\cdot|x'_i), ~\delta_{y'_i}\right),}
\end{align}
where $\psi,\theta$ are the network parameters we wish to learn. In discrminative learning applications, the factor $\pi_\theta(\cdot|x)$ can be readily used as a probabilistic classifier (e.g., section~\ref{sec:prompt}).

\subsubsection{Transport Plan with Implicit Models}\label{imp}

As mentioned earlier, in applications such as {$\S$~\ref{sec:simbary}, ~$\S$~\ref{sec:simbio}}, it is required to generate samples from $\pi_{Y'|Y,X}(\cdot|y,x)$ for inference. In such applications, one would prefer modelling these transport plan factors using implicit generative models.
 
\begin{figure}[t]
\begin{center}
\includegraphics[width=0.7\columnwidth]{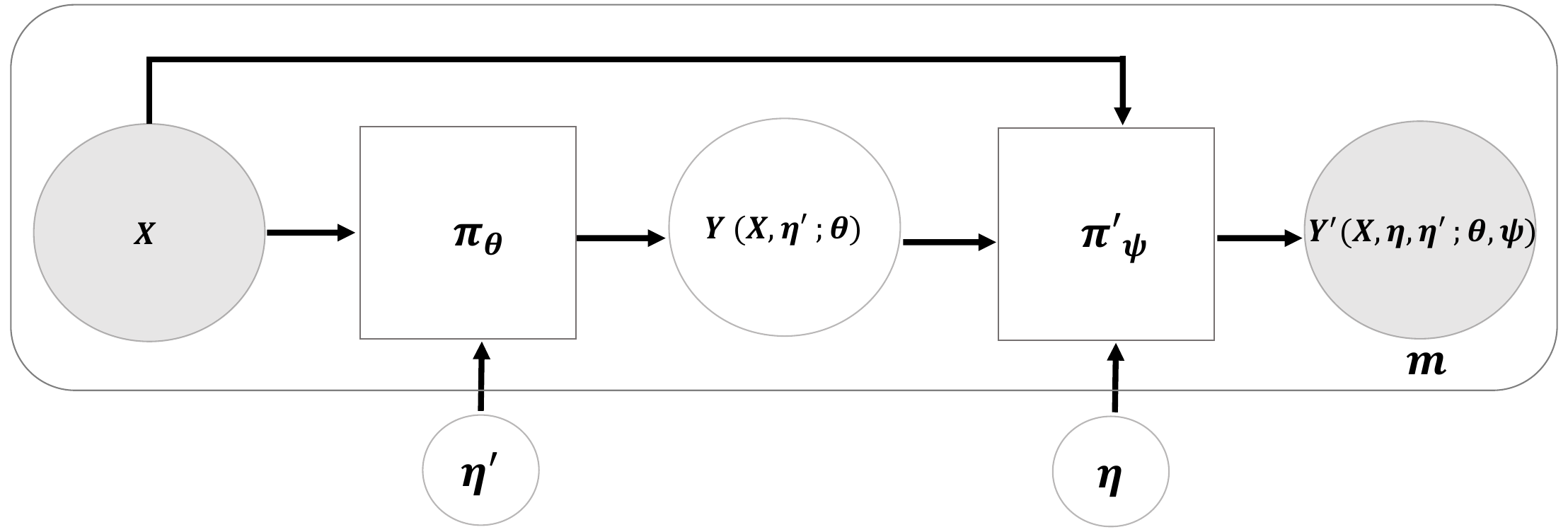}
\caption{Illustration of the proposed factorization and implicit modelling for learning the transport plan $\pi_{Y, Y'|X}(y, y'|x)$ through the factors $\pi_\theta(y|x) \pi_\psi(y'|y, x)$, parameterized by fixed-architecture neural networks \ref{imp}. $\eta, \ \eta'\sim \mathcal{N}(0, 1)$ denotes the noise input to the implicit models.}
\label{implicit-model}
\end{center}
\end{figure}

Since the MMD metric, unlike KL-divergence, can be employed to compare measures with non-overlapping support, implicit generative models can be readily employed for modelling our transport plan. More specifically, we model the factors $\pi_{Y'|Y, X}(y'|y, x)$, $\pi_{Y|X}(y|x)$ with fixed-architecture generative neural networks, $\pi_\psi$ and $\pi_\theta$, respectively. We use $\eta,~\eta'\sim \mathcal{N}(0, 1)$ to denote the noise random variables. The $\pi_\theta$ network takes as input $x$ and random $\eta'$ to produce (random) $y$, to be distributed as $\pi_{Y|X}(\cdot|x)$. Like-wise, the $\pi_\psi$ network takes as input $y,x$ and random $\eta$ to produce (random) $y'$, to be distributed as $\pi_{Y'|Y,X}(\cdot|y,x)$. We denote the outputs of $\pi_\theta$ by $y(x, \eta'_i;\theta)\ i=1,\ldots,m$ (i.e., samples from $\pi_{Y/X}(\cdot|x)$. And, we denote outputs of $\pi_\psi$ by $y\left(x, \eta_i, \eta'_i;\theta,\psi\right)\ i=1,\ldots,m$, when inputs  are $y(x, \eta'_i;\theta),x,\eta_i$. We illustrate the overall model in figure~\ref{implicit-model}. Then, the COT estimator, with implicit modelling, reads as:
\begin{align}\label{eqn:impcot}
&\scriptstyle{\min\limits_{\theta, \psi}\int_{\calX}\frac{1}{m}\sum_{i=1}^m c\left(y(x, \eta'_i;\theta), y\left(x, \eta_i, \eta_i';\theta,\psi\right)\right)\textup{d}a}(x) \nonumber \\&\qquad\scriptstyle{+\lambda_1\frac{1}{m}\sum_{i=1}^m\MMD^2\left(\frac{1}{m}\sum_{j=1}^m\delta_{y\left(x_i, \eta_j, \eta'_j;\theta,\psi\right) }, \delta_{y_i}\right)} \nonumber \\&\qquad\scriptstyle{+\lambda_2\frac{1}{m}\sum_{i=1}^m\MMD^2\left(\frac{1}{m}\sum_{j=1}^m\delta_{y(x'_i, \eta'_j;\theta) }, \delta_{y'_i}\right).}
\end{align}

We note that solving the COT problem, then readily provides us with the factors $\pi_{Y'|Y,X}(y'|y,x)$ and $\pi_{Y|X}(y|x)$, which can be used for inference purposes. This is in contrast to a typical implicit modelling approach, where one would require samples of $(x,y,y')$ for learning such a model. The unavailability of such triplets (as in {~$\S$~\ref{sec:simbio}}) often limits such typical approaches. However, as we can see, COT now allows us to learn such a model without the availability of such triplets, only using samples from $s_{X,Y}$ and $t_{X', Y'}$. This clearly shows the benefits of the proposed approach.

\section{EXPERIMENTS}\label{sec:exp}
In this section, we showcase the utility of the proposed estimator \ref{eqn:regcot2} in various applications. 
We choose the auxiliary distribution $a$ as the empirical distribution over the training covariates and use $\lambda_1=\lambda_2=\lambda$ in all our experiments. More experimental details and results are in Supplementary $\S~(\textup{S}2.2)$. \footnote{The code for reproducing our experiments is publicly available at \texttt{https://github.com/atmlr-lab/COT}.}
\subsection{Verifying Correctness of Estimator}
We empirically verify the correctness of the proposed estimator in synthetically constructed settings where the closed-form solutions are known.
\subsubsection{Convergence to the True Wasserstein}
We learn the implicit networks with the proposed COT loss~\ref{eqn:impcot}, keeping $\lambda$ high enough. With the learnt networks, we draw samples $y(x, \eta'_i; \theta)\sim \pi_\theta(\cdot|x)$ and $y(x, \eta_i, \eta'_i; \theta, \psi)\sim \pi_\psi(\cdot|y(x, \eta'_i; \theta), x)$ , for $i=1,\cdots, m$, and compute the transport cost (first term in \ref{eqn:impcot}) and compare it with ${W}_c(s_{Y|X}(\cdot|x),t_{Y'|X'}(\cdot|x))$. In order to verify that our estimate converges to the true Wasserstein, we consider a case where the analytical solution for the Wasserstein distance $W_c$ is known and compare it with our estimate.
\paragraph{Experimental Setup} We consider two distributions $y \sim \mathcal{N}(4(x-0.5),1)$ and $y' \sim \mathcal{N}(-2(x'-0.5),8x'+1)$ where $x \sim \beta(2,4)$ and $x \sim \beta(4,2)$ generate $m$ samples from each them. The true Wasserstein distance between them at $x$ turns out to be $(6(x-0.5))^2 + (\sqrt{8x+1}-1)^2$ (see Equation (2.39) in \cite{CompOT}), which we compare against. We use the RBF kernel and squared Euclidean distance as our ground cost. The factors $\pi_\theta(\cdot|x)$ and $\pi_\psi(\cdot|y,x)$ are modelled using two 2-layer MLP neural networks.

\begin{figure}[t]
    \centering
    \includegraphics[width=\columnwidth]{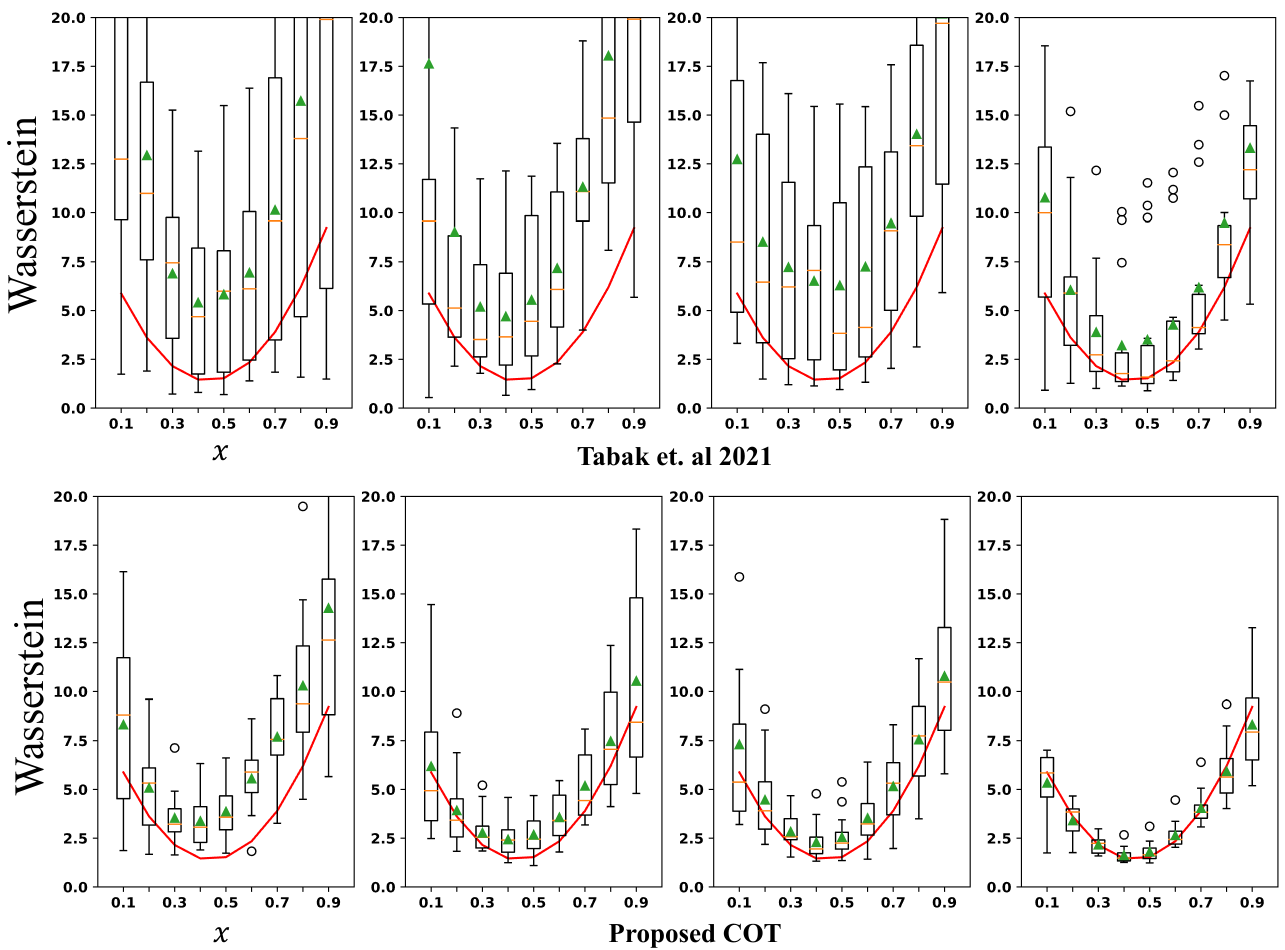}
    \caption{As $m\in\{100, 200, 400, 800\}$ increases from left to right, we plot the true Wasserstein distance in red and mark the means (in orange) and medians (in green) of the distances estimated using~\cite{Tabak21} and the proposed COT estimator. The statistics are obtained from runs over multiple seeds. The corresponding MSEs are $\{ 245.530, 290.458, 89.715, 27.687\}$ and $\{22.711, 6.725, 8.052, 1.580\}$ respectively. It can be seen that the proposed COT objective converges to the true Wasserstein faster than~\cite{Tabak21}.}
    \label{'fig:correctness'}
\end{figure}
\paragraph{Results} Figure~\ref{'fig:correctness'} shows the convergence to the true Wasserstein as $m$ increases. The variance of the estimated values and the MSEs decrease as the number of samples increases. The quadratic nature of the function is also captured with our estimator.
\subsubsection{Convergence to the True Barycenter}\label{sec:simbary}
\begin{figure}[t]
    \centering
    \includegraphics[width=0.9\columnwidth]{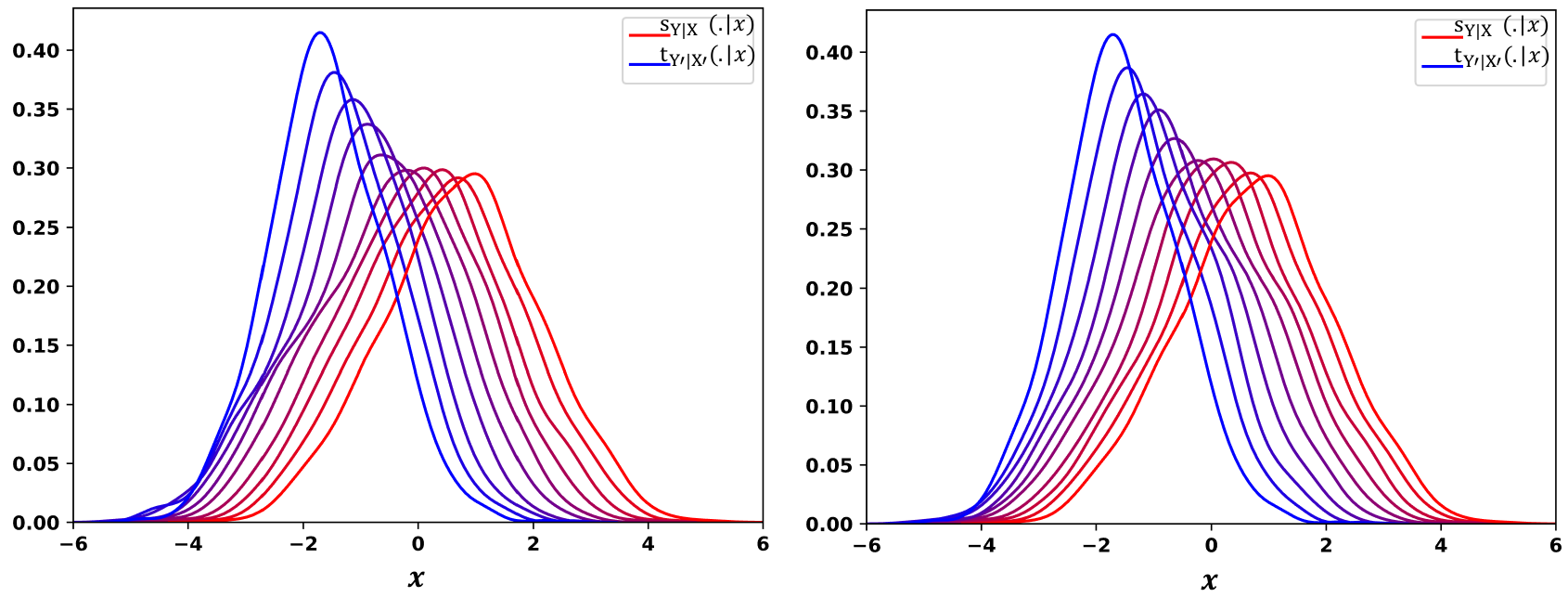}
    \caption{Barycenters shown on varying $\rho\in[0, 1]$ with colors interpolated between red and blue. Left: Conditional barycenter learnt by the proposed COT method. Right: Analytical barycenter.}
    \label{fig:my_label11}
\end{figure}

\begin{figure}[t]
    \centering
    \includegraphics[width=0.28\textwidth]{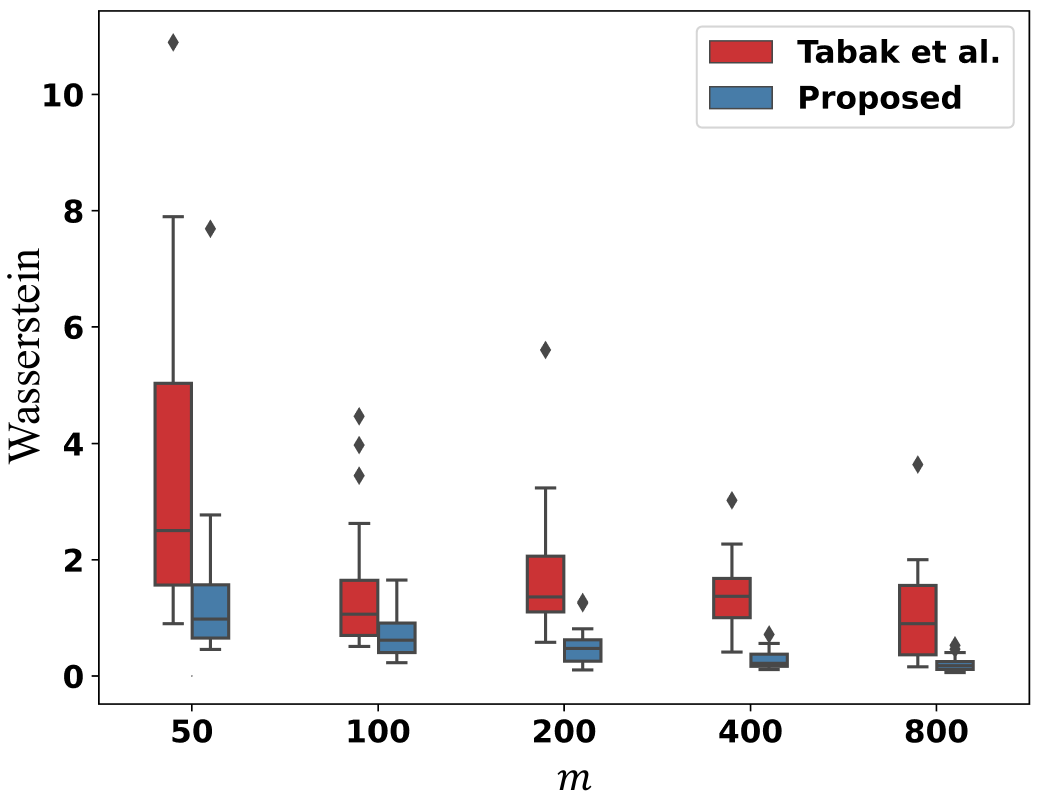}
    \caption{For increasing values of $m$, we show box plots of the Wasserstein distance between the learnt barycenter, $B_x$, and the analytical barycenter. The corresponding MSEs are $\{22.399,  3.408, 3.964, 2.534, 1.687\}$ for ~\cite{Tabak21} and $\{4.441, 0.654, 0.353, 0.099, 0.058\}$ for the proposed COT estimator. It can be seen that the proposed COT-based barycenter converges to the true solution faster than~\cite{Tabak21}.}
    \label{fig:my_label10}
\end{figure}

For further verification of our estimator, we show that the barycenter estimated using our transport plan and the true barycenter converge in Wasserstein distance. 
\paragraph{Experimental Setup} Two independent Gaussian distributions are taken $y \sim \mathcal{N}(2(x-0.5),1)$ and $y' \sim \mathcal{N}(-4(x'-0.5),4)$ where  $x \sim \beta(2,4)$ and $x' \sim \beta(4,2)$.  The analytical solution of the barycenter is calculated as $y_c \sim \mathcal{N}(-x+0.5,2.5)$ \cite{CompOT}. 
Recall that the barycenter can also be computed using the optimal transport map (Remark 3.1 in \cite{pmlr-v97-gordaliza19a}) using the expression: $B_x= \rho S_x+(1-\rho)T_x,$ where $\rho\in[0, 1]$ and  $B_x,~S_x, ~T_x$ denote the random variables corresponding to the barycenter, source measure and the transported sample, conditioned on $x$, respectively. 
Accordingly, samples from the barycenter, $B_{x_i}$, are obtained using: $\rho y_i+ (1-\rho)y$, where $y \sim \pi_\psi(\cdot|y_i, x_i)$.
\paragraph{Results} For evaluation, we generate 500 samples from our transport plan based barycenter and the true barycenter. We use kernel density estimation to plot the barycenters. Figures~\ref{fig:my_label11} and ~\ref{fig:my_label10} show that the proposed estimate of barycenter closely resembles the analytical barycenter and converges on increasing $m$.

\subsection{Cell Population Dynamics}
\label{sec:simbio}
The study of single-cell molecular responses to treatment drugs is a major problem in biology. Existing single-cell-sequencing methods allow one to observe gene expressions of the cells, but do so by destroying them. As a result, one ends up with cells from control (unperturbed) and target (perturbed) distributions without a correspondence between them. Optimal transport has emerged as a natural method \cite{bunne2021learning} to obtain a mapping between the source and the target cells, which can then be used for predictions on unseen cells. As the drug dosage is highly correlated with the predicted cell populations, \cite{Cuturi22} learns such optimal transport maps conditioned on the drug dosage. We apply the proposed COT formulation to generate samples from the distributions over perturbed cells conditioned on the drug dosage given to an unperturbed cell.




\paragraph{Dataset} We consider the dataset used by \cite{Cuturi22} and \cite{bunne2021learning} corresponding to the cancer drug \texttt{Givinostat} applied at different dosage levels, $\{x_1=10nM, x_2=100nM, x_3=1000nM, x_4=10000nM\}$. At each dosage level, $x_i$, samples of perturbed cells are given: $y_{i1},\ldots,y_{im_i}$. The total perturbed cells are 3541. Samples of unperturbed cells are also provided: $y'_1,\ldots,y'_m, m=17,565$. Each of these cells is described by gene-expression levels of $n=1000$ highly variable genes, i.e., $y_{ij},y'_i\in\R^{1000}$. Following \cite{Cuturi22}, the representations of cells are brought down to 50 dimensions with PCA.

\paragraph{COT-Based Generative Modelling} Our goal is to perform OT between the distribution of the unperturbed cells and the distribution of the perturbed cell conditioned on the drug dosage. As the representations of the cells lie in $\calY=\R^{50}$, we choose implicit modelling ($\S$~\ref{imp}) for learning the conditional transport plans. The factor $\pi_\theta$ is taken as the empirical distribution over the unperturbed cells. With this notation, our COT estimator, (\ref{eqn:impcot}), simplifies as follows.
\begin{align*}
&\scriptstyle{\min\limits_{\psi}\frac{1}{4}\sum_{q=1}^4\frac{1}{m}\sum_{i=1}^{m} c\left(y'_{i}, y\left(x_q,\eta_{i};\psi\right)\right)} \\&\qquad\scriptstyle{+\lambda_1\frac{1}{4}\sum_{i=1}^4\MMD^2\left(\frac{1}{m}\sum_{j=1}^{m}\delta_{y\left(x_i, \eta_{j};\psi\right) }, \frac{1}{m_i}\sum_{j=1}^{m_i}\delta_{y_{ij}}\right),}
\end{align*}
where $y\left(x, \eta_{i};\psi\right)\ i=1,\ldots,m$ are samples from the network $\pi_\psi(\cdot|y'_{i},x)$.

\paragraph{Experimental Setup}



Similar to \cite{Cuturi22}, we take the cost function, $c$, as squared Euclidean. For the MMD regularization, we use the characteristic inverse multi-quadratic (IMQ) kernel. 

\paragraph{Results} 

Following \cite{Cuturi22}, we evaluate the performance of COT by comparing samples from the predicted and ground truth perturbed distributions. We report the $l_2$ norm between the Perturbation Signatures \cite{Stathias2018DrugAD}, for 50 marker genes for various dosage levels. We also report the MMD distances between the predicted and target distributions on various dosage levels. The distances are reported for in-sample settings, i.e. the dosage levels are seen during training. We compare our performance to the reproduced CellOT~\cite{bunne2021learning} and CondOT \cite{Cuturi22} baselines.

We summarize our results in Tables \ref{sample-table-l2} and \ref{sample-table-mmd}. We observe that COT consistently outperforms state-of-the-art baselines CondOT~\cite{Cuturi22} and CellOT~\cite{bunne2021learning} in terms of $l_2$ (PS) as well as the MMD distances.


    

\begin{table}[t]
  \caption{$l_2$ (PS) distances (lower is better) between predicted and ground truth distributions}
  \label{sample-table-l2}
  \centering
  \footnotesize{
  \begin{tabular}{lccc}
    \toprule
        Dosage   & CellOT & CondOT & Proposed \\  
        \midrule
        $10nM$ & 1.2282 & 0.3789 & \textbf{0.3046} \\
        $100nM$ & 1.2708 & 0.2515 & \textbf{0.2421}\\
        $1000nM$ & 0.8653 & 0.7290 & \textbf{0.3647} \\
        $10000nM$ & 4.9035 & 0.3819 & \textbf{0.2607}  \\
        \textbf{Average} & 2.067 & 0.4353 &  \textbf{0.2930}\\
        \bottomrule 
  \end{tabular}}
\end{table}

\begin{table}[t]
  \caption{MMD distances (lower is better) between predicted and ground truth distributions}
  \label{sample-table-mmd}
  \centering
  \footnotesize{
  \begin{tabular}{lccc}
    \toprule
        Dosage  & CellOT & CondOT & Proposed \\  
        \midrule
        $10nM$ & 0.01811 &  0.00654 & \textbf{0.00577} \\
        $100nM$ & 0.0170 & 0.00555 & \textbf{0.00464}\\
        $1000nM$ & 0.0154 & 0.01290 & \textbf{0.00647} \\
        $10000nM$ & 0.1602 & 0.01034 & \textbf{0.00840}  \\
        \textbf{Average} & 0.0526 & 0.00883 & \textbf{0.00632} \\
        \bottomrule 
  \end{tabular}}
\end{table}

\subsection{Prompt Learning}\label{sec:prompt}
\begin{figure}[t]
    \centering
    \includegraphics[width=0.9\columnwidth]{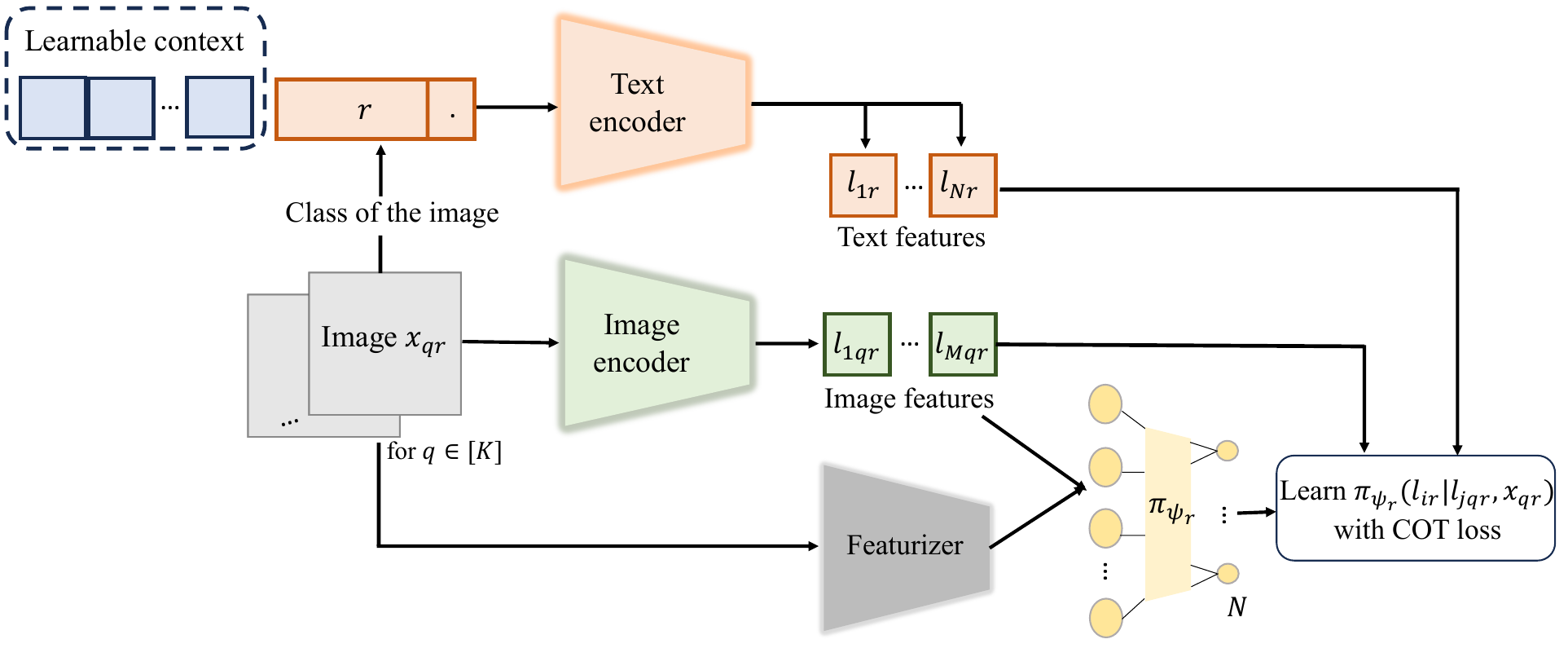}
    \caption{We pose learning prompts in few-shot classification as the conditional optimal transport problem. The figure shows our neural network diagram for learning conditional optimal transport plans.}
    \label{prompt-diag}
\end{figure}
In order to show the versatility of our framework, we adapt our estimator for learning prompts for large-scale vision-language models and evaluate the performance in a limited supervision setting. 

The success of vision-language models in open-world visual understanding has motivated efforts which aim to learn prompts \cite{zhou2022cocoop, zhang2021tip, coop,chen2023plot} to adapt the knowledge from pre-trained models like CLIP~\cite{CLIP} for downstream tasks since it is infeasible to fine-tune such models due to a large number of parameters. Typically, these approaches rely on learning class-specific prompts for each category to better adapt the vision-language model for downstream tasks without the need for fine-tuning. A recent approach, PLOT \cite{chen2023plot}, achieved state-of-the-art results by incorporating an OT-based loss between distributions over the set of local visual features and the set of textual prompt features, each of 1024 dimensions, to learn the downstream classifier. For each image, PLOT computes an OT-based loss between $M (49)$ visual features of the image and $N (4)$ textual prompt features per class. 


 As prompts are shared across images of a class~\cite{chen2023plot}, learning optimal transport plans conditioned on class-level information is expected to improve the downstream performance compared to solving an OT problem separately for each (image, class) pair. Hence, we pose this prompt learning task as a COT problem, where the conditional transport plans are modelled explicitly ($\S$~\ref{expl}).
\paragraph{Validating the Proposed Explicit Modelling}
Before working on the challenging few-shot classification task, we evaluate the proposed explicit modelling-based COT estimator on a simpler multi-class classification task. Let the discriminative model to be  learnt be $f_\theta$. The idea is to match this conditional to that in the training data using COT. We choose the transport plan factor $\pi_\theta\equiv f_\theta$ and $a$ as the marginal of input covariates in the training data, simplifying our COT estimator, (\ref{eqn:expexpcot}), as:
\begin{align}\nonumber
\scriptstyle{\min\limits_{\psi,\theta}}&\scriptstyle{\frac{1}{m}\sum_{q=1}^m\sum_{i=1,j=1}^{i=n,j=n} c(l_i,l_j)\pi_\psi(l_i|l_j,x_q)f_\theta(l_j|x_q)} \nonumber\\& \scriptstyle{+\lambda_1\frac{1}{m}\sum_{i=1}^m\MMD^2\left(\sum_{j=1}^n\pi_\psi(\cdot|l_j,x_i)f_\theta(l_j|x_i), \delta_{y_i}\right),}
\end{align}
where $\psi,\theta$ are the network parameters we wish to learn. 
Table~\ref{tab:classifier} validates the performance with the proposed explicit modelling.

\begin{table}[t]
\caption{AUC on test data (higher is better). We compare the performance of COT against other OT-based losses $\epsilon$-OT (\cite{Frogner15}) and CKB (\cite{Bures}).}
\begin{center}
\footnotesize{\begin{tabular}{cccc}
        \toprule
        Dataset & $\epsilon$-OT & CKB & Proposed \\
        \midrule
        MNIST & 0.89 &  \textbf{0.99}  & \textbf{0.99}\\
        CIFAR10 & 0.66 & 0.73 & \textbf{0.79}\\
        Animals with Attribute & 0.68 & 0.64 & \textbf{0.86}\\
        \bottomrule
      \end{tabular}}
      \label{tab:classifier}
\end{center}
\end{table}

\paragraph{COT Formulation for Prompt Learning}
We learn an explicit model $\pi_{\psi_r}(\cdot|l_{jqr},x_{qr})$ over the $N$ textual prompt features $l_{1r},\ldots,l_{Nr}$ for each class. Here, $x_{qr}$ is the $q^{th}$ image from class $r$ and $l_{jqr}$ is the $j^{th}$ visual feature for image $x_{qr}$. Following PLOT, the distribution over image features given an image is considered uniform and, hence, not modelled as the other factor in the transport plan. Figure~(\ref{prompt-diag}) depicts the proposed setup. Our formulation for prompt learning for $K$-shot classification (only $K$ training images per class) is as follows.
\begin{align}\nonumber
\scriptstyle{\min\limits_{\psi_r}}&\scriptstyle{\frac{1}{K}\sum_{q=1}^K\sum_{i=1,j=1}^{i=N,j=M} c(l_{ir},l_{jqr})\pi_{\psi_r}(l_{ir}|l_{jqr},x_{qr}) \mathbf{v}_j}\\&\scriptstyle{~+\lambda_1\MMD^2\left(\sum_{q=1}^K\sum_{j=1}^{M}\pi_{\psi_r}(\cdot|l_{jqr},x_{qr})\mathbf{v}_j, \mathbf{u}\right).}\label{cot-prompt}
 \end{align}
Following the PLOT setup, we take $\mathbf{v}, \mathbf{u}$ as uniform distributions over the $M(49)$ visual features and the $N(4)$ prompt features, respectively. As the prompts are shared across the images of a class, our MMD-regularization term matches the cumulative marginals to the distribution over prompt features.
\paragraph{Experimental Setup}
We take the same experimental setup used in CoOp\cite{coop} and PLOT\cite{chen2023plot} for learning prompts and only change the training loss to \ref{cot-prompt}. The kernel employed is the characteristic inverse multi-quadratic, and the ground cost is the cosine cost. We follow the common training/evaluation protocol used in CoOp and PLOT and report the mean and standard deviation of the accuracies obtained with 3 seeds.

\textbf{Results} In Table \ref{table:prompt}, we report the accuracies on the EuroSAT benchmark dataset \cite{helber2019eurosat} for the number of shots $K$ as 1, 2, 4 and 8. As the number of shots represents the number of training images per class, learning with lesser $K$ is more difficult. The advantage of class-level context brought by the proposed COT formulation is evident in this setting. 

\begin{table}[t]
  \caption{Prompt Learning experiment: Average accuracy (higher is better) on EuroSAT dataset. The class-level context brought by the proposed COT method allows it to outperform the state-of-the-art PLOT baseline, especially in the challenging case of lesser $K$.}
  \label{table:prompt}
  \centering
  \footnotesize{
  \begin{tabular}{lccc}
    \toprule
        &  CoOp  & PLOT & Proposed \\  
        \midrule
        $K=1$ & 52.12 $\pm$ 5.46 & 54.05 $\pm$ 5.95 & \textbf{61.20 $\pm$ 3.65}\\
        $K=2$ & 59.00 $\pm$ 3.48 & 64.21 $\pm$ 1.90 & \textbf{64.67 $\pm$ 2.37}\\
        $K=4$ & 68.61 $\pm$ 3.54 & 72.36 $\pm$ 2.29 & \textbf{72.53 $\pm$ 2.60}\\
        $K=8$ & 77.08 $\pm$ 2.42 & 78.15 $\pm$ 2.65 & \textbf{78.57 $\pm$ 2.38}\\
        \bottomrule 
  \end{tabular}}
\end{table}

\section{CONCLUSION}
Often, machine learning applications need to compare conditional distributions. Remarkably, our framework enables such a comparison solely using samples from (observational) joint distributions. To the best of our knowledge, the proposed method is the first work that consistently estimates the conditional transport plan in the general setting. The cornerstone of our work lies in the theoretical analysis of its convergence properties, demonstrating different modelling choices for learning and empirically validating its correctness. We further showcase the utility of the proposed method in downstream applications of cell population dynamics and prompt learning for few-shot classification. A possible future work would be to extend the proposed approach of generating conditional barycenters ($\S$~\ref{sec:simbary}) to work with more than two conditionals.



\subsubsection*{Acknowledgements}
The first author is supported by the Google PhD Fellowship. JSN would like to thank Fujitsu Limited, Japan, for the generous research grant. We thank Charlotte Bunne for the clarifying discussions on reproducing the CondOT method. We also thank Dr Pratik Jawanpuria, Kusampudi Venkata Datta Sri Harsha, Shivam Chandhok, Aditya Saibewar, Amit Chandhak and the anonymous reviewers who helped us improve our work. PM thanks Suvodip Dey and Sai Srinivas Kancheti for the support.

\bibliographystyle{apalike}
\bibliography{example_paper}

\begin{thebibliography}{}

\bibitem[Bojanowski et~al., 2017]{bojanowski-etal-2017-enriching}
Bojanowski, P., Grave, E., Joulin, A., and Mikolov, T. (2017).
\newblock Enriching word vectors with subword information.
\newblock {\em Transactions of the Association for Computational Linguistics}, 5:135--146.

\bibitem[Bunne et~al., 2022]{Cuturi22}
Bunne, C., Krause, A., and Cuturi, M. (2022).
\newblock Supervised training of conditional monge maps.
\newblock In {\em NeurIPS}.

\bibitem[Bunne et~al., 2021]{bunne2021learning}
Bunne, C., Stark, S.~G., Gut, G., del Castillo, J.~S., Lehmann, K.-V., Pelkmans, L., Krause, A., and R{\"a}tsch, G. (2021).
\newblock Learning single-cell perturbation responses using neural optimal transport.
\newblock {\em bioRxiv}.

\bibitem[Bunne et~al., 2023]{cellot}
Bunne, C., Stark, S.~G., Gut, G., del Castillo, J.~S., Levesque, M., Lehmann, K.-V., Pelkmans, L., Krause, A., and Ratsch, G. (2023).
\newblock Learning single-cell perturbation responses using neural optimal transport.
\newblock {\em Nature Methods}.

\bibitem[Cao et~al., 2022]{Cao22}
Cao, Z., Xu, Q., Yang, Z., He, Y., Cao, X., and Huang, Q. (2022).
\newblock Otkge: Multi-modal knowledge graph embeddings via optimal transport.
\newblock In {\em NeurIPS}.

\bibitem[Chen et~al., 2023]{chen2023plot}
Chen, G., Yao, W., Song, X., Li, X., Rao, Y., and Zhang, K. (2023).
\newblock Prompt learning with optimal transport for vision-language models.
\newblock In {\em ICLR}.

\bibitem[Fatras et~al., 2021]{fatras2021jumbot}
Fatras, K., S\'ejourn\'e, T., Courty, N., and Flamary, R. (2021).
\newblock Unbalanced minibatch optimal transport; applications to domain adaptation.
\newblock In {\em ICML}.

\bibitem[Fatras et~al., 2020]{fatras2019learnwass}
Fatras, K., Zine, Y., Flamary, R., Gribonval, R., and Courty, N. (2020).
\newblock Learning with minibatch wasserstein: asymptotic and gradient properties.
\newblock In {\em AISTATS}.

\bibitem[Frogner et~al., 2015]{Frogner15}
Frogner, C., Zhang, C., Mobahi, H., Araya, M., and Poggio, T.~A. (2015).
\newblock Learning with a wasserstein loss.
\newblock In {\em NIPS}.

\bibitem[Gordaliza et~al., 2019]{pmlr-v97-gordaliza19a}
Gordaliza, P., Barrio, E.~D., Fabrice, G., and Loubes, J.-M. (2019).
\newblock Obtaining fairness using optimal transport theory.
\newblock In Chaudhuri, K. and Salakhutdinov, R., editors, {\em Proceedings of the 36th International Conference on Machine Learning}, volume~97 of {\em Proceedings of Machine Learning Research}, pages 2357--2365. PMLR.

\bibitem[Graham et~al., 2020]{Graham2020MinimaxRA}
Graham, B.~S., Niu, F., and Powell, J.~L. (2020).
\newblock Minimax risk and uniform convergence rates for nonparametric dyadic regression.
\newblock {\em NBER Working Paper Series}.

\bibitem[Gr{\"{u}}new{\"{a}}lder et~al., 2012]{gretton}
Gr{\"{u}}new{\"{a}}lder, S., Lever, G., Gretton, A., Baldassarre, L., Patterson, S., and Pontil, M. (2012).
\newblock Conditional mean embeddings as regressors.
\newblock In {\em ICML}.

\bibitem[Hahn et~al., 2019]{hahn2019atlantic}
Hahn, P.~R., Dorie, V., and Murray, J.~S. (2019).
\newblock Atlantic causal inference conference ({ACIC}) data analysis challenge 2017.

\bibitem[Helber et~al., 2019]{helber2019eurosat}
Helber, P., Bischke, B., Dengel, A., and Borth, D. (2019).
\newblock Eurosat: A novel dataset and deep learning benchmark for land use and land cover classification.
\newblock {\em IEEE Journal of Selected Topics in Applied Earth Observations and Remote Sensing}, 12(7):2217--2226.

\bibitem[Jawanpuria et~al., 2021]{drpratikrot4c}
Jawanpuria, P., Satyadev, N., and Mishra, B. (2021).
\newblock Efficient robust optimal transport with application to multi-label classification.
\newblock In {\em IEEE Conference on Decision and Control (CDC)}.

\bibitem[Kantorovich, 1942]{KatoroOT}
Kantorovich, L. (1942).
\newblock On the transfer of masses (in russian).
\newblock {\em Doklady Akademii Nauk}, 37(2):227--229.

\bibitem[Kidger and Lyons, 2020]{pmlr-v125-kidger20a}
Kidger, P. and Lyons, T. (2020).
\newblock {Universal Approximation with Deep Narrow Networks}.
\newblock In {\em ICML}.

\bibitem[Krizhevsky et~al., 2009]{krizhevsky2009learning}
Krizhevsky, A., Hinton, G., et~al. (2009).
\newblock Learning multiple layers of features from tiny images.

\bibitem[Lampert et~al., 2009]{5206594}
Lampert, C.~H., Nickisch, H., and Harmeling, S. (2009).
\newblock Learning to detect unseen object classes by between-class attribute transfer.
\newblock In {\em CVPR}.

\bibitem[LeCun and Cortes, 2010]{mnist}
LeCun, Y. and Cortes, C. (2010).
\newblock {MNIST} handwritten digit database.

\bibitem[Li et~al., 2022]{LiNeykovBalakrishnan}
Li, M., Neykov, M., and Balakrishnan, S. (2022).
\newblock {Minimax optimal conditional density estimation under total variation smoothness}.
\newblock {\em Electronic Journal of Statistics}, 16(2):3937 -- 3972.

\bibitem[Liu et~al., 2021]{Liu2021WassersteinGL}
Liu, S., Zhou, X., Jiao, Y., and Huang, J. (2021).
\newblock Wasserstein generative learning of conditional distribution.
\newblock {\em ArXiv}.

\bibitem[Liu et~al., 2020]{liu2020semantic}
Liu, Y., Zhu, L., Yamada, M., and Yang, Y. (2020).
\newblock Semantic correspondence as an optimal transport problem.
\newblock In {\em CVPR}.

\bibitem[Luo and Ren, 2021]{Bures}
Luo, Y.-W. and Ren, C.-X. (2021).
\newblock Conditional bures metric for domain adaptation.
\newblock In {\em CVPR}.

\bibitem[Maurer, 2016]{rad}
Maurer, A. (2016).
\newblock A vector-contraction inequality for rademacher complexities.
\newblock In {\em ALT}.

\bibitem[Muandet et~al., 2017]{MAL-060}
Muandet, K., Fukumizu, K., Sriperumbudur, B., and Schölkopf, B. (2017).
\newblock Kernel mean embedding of distributions: A review and beyond.
\newblock {\em Foundations and Trends® in Machine Learning}, 10(1-2):1--141.

\bibitem[Neyshabur, 2017]{neyshabur2017implicit}
Neyshabur, B. (2017).
\newblock Implicit regularization in deep learning.

\bibitem[Peyré and Cuturi, 2019]{CompOT}
Peyré, G. and Cuturi, M. (2019).
\newblock Computational optimal transport.
\newblock {\em Foundations and Trends® in Machine Learning}, 11(5-6):355--607.

\bibitem[Radford et~al., 2021]{CLIP}
Radford, A., Kim, J.~W., Hallacy, C., Ramesh, A., Goh, G., Agarwal, S., Sastry, G., Askell, A., Mishkin, P., Clark, J., Krueger, G., and Sutskever, I. (2021).
\newblock Learning transferable visual models from natural language supervision.
\newblock In {\em ICML}.

\bibitem[Song et~al., 2009]{Song2009HilbertSE}
Song, L., Huang, J., Smola, A., and Fukumizu, K. (2009).
\newblock Hilbert space embeddings of conditional distributions with applications to dynamical systems.
\newblock In {\em ICML}.

\bibitem[Sriperumbudur et~al., 2011]{characteristicK}
Sriperumbudur, B.~K., Fukumizu, K., and Lanckriet, G. R.~G. (2011).
\newblock Universality, characteristic kernels and {RKHS} embedding of measures.
\newblock {\em Journal of Machine Learning Research}, 12:2389–2410.

\bibitem[Stathias et~al., 2018]{Stathias2018DrugAD}
Stathias, V., Jermakowicz, A.~M., Maloof, M.~E., Forlin, M., Walters, W.~M., Suter, R.~K., Durante, M.~A., Williams, S.~L., Harbour, J.~W., Volmar, C.-H., Lyons, N.~J., Wahlestedt, C., Graham, R.~M., Ivan, M.~E., Komotar, R.~J., Sarkaria, J.~N., Subramanian, A., Golub, T.~R., Sch{\"u}rer, S.~C., and Ayad, N.~G. (2018).
\newblock Drug and disease signature integration identifies synergistic combinations in glioblastoma.
\newblock {\em Nature Communications}, 9.

\bibitem[Séjourné et~al., 2023a]{sliced-uot}
Séjourné, T., Bonet, C., Fatras, K., Nadjahi, K., and Courty, N. (2023a).
\newblock Unbalanced optimal transport meets sliced-wasserstein.

\bibitem[Séjourné et~al., 2023b]{sduot}
Séjourné, T., Feydy, J., Vialard, F.-X., Trouvé, A., and Peyré, G. (2023b).
\newblock Sinkhorn divergences for unbalanced optimal transport.

\bibitem[Tabak et~al., 2021]{Tabak21}
Tabak, E.~G., Trigila, G., and Zhao, W. (2021).
\newblock Data driven conditional optimal transport.
\newblock {\em Machine Learning}, 110(11):3135--3155.

\bibitem[Waarde and Sepulchre, 2022]{pmlr-v168-waarde22a}
Waarde, H.~v. and Sepulchre, R. (2022).
\newblock Training lipschitz continuous operators using reproducing kernels.
\newblock In {\em Annual Learning for Dynamics and Control Conference}.

\bibitem[Wolf et~al., 2018]{Wolf2018}
Wolf, F.~A., Angerer, P., and Theis, F.~J. (2018).
\newblock Scanpy: large-scale single-cell gene expression data analysis.
\newblock {\em Genome Biology}, 19(1):15.

\bibitem[Zhang et~al., 2022]{zhang2021tip}
Zhang, R., Zhang, W., Fang, R., Gao, P., Li, K., Dai, J., Qiao, Y., and Li, H. (2022).
\newblock Tip-{A}dapter: Training-free adaption of clip for few-shot classification.
\newblock In {\em ECCV}.

\bibitem[Zhou et~al., 2022a]{zhou2022cocoop}
Zhou, K., Yang, J., Loy, C.~C., and Liu, Z. (2022a).
\newblock Conditional prompt learning for vision-language models.
\newblock In {\em CVPR}.

\bibitem[Zhou et~al., 2022b]{coop}
Zhou, K., Yang, J., Loy, C.~C., and Liu, Z. (2022b).
\newblock Learning to prompt for vision-language models.
\newblock {\em International Journal of Computer Vision}, 130(9):2337–2348.

\end{thebibliography}
\section*{Checklist}


 \begin{enumerate}

 \item For all models and algorithms presented, check if you include:
 \begin{enumerate}
   \item A clear description of the mathematical setting, assumptions, algorithm, and/or model. [Yes]
   \item An analysis of the properties and complexity (time, space, sample size) of any algorithm. [Yes]
   \item (Optional) Anonymized source code, with specification of all dependencies, including external libraries. [Yes]
 \end{enumerate}

 \item For any theoretical claim, check if you include:
 \begin{enumerate}
   \item Statements of the full set of assumptions of all theoretical results. [Yes]
   \item Complete proofs of all theoretical results. [Yes]
   \item Clear explanations of any assumptions. [Yes]     
 \end{enumerate}

 \item For all figures and tables that present empirical results, check if you include:
 \begin{enumerate}
   \item The code, data, and instructions needed to reproduce the main experimental results (either in the supplemental material or as a URL). [Yes]
   \item All the training details (e.g., data splits, hyperparameters, how they were chosen). [Yes]
         \item A clear definition of the specific measure or statistics and error bars (e.g., with respect to the random seed after running experiments multiple times). [Yes]
         \item A description of the computing infrastructure used. (e.g., type of GPUs, internal cluster, or cloud provider). [Yes]
 \end{enumerate}

 \item If you are using existing assets (e.g., code, data, models) or curating/releasing new assets, check if you include:
 \begin{enumerate}
   \item Citations of the creator If your work uses existing assets. [Yes]
   \item The license information of the assets, if applicable. [Not Applicable]
   \item New assets either in the supplemental material or as a URL, if applicable. [Yes]
   \item Information about consent from data providers/curators. [Not Applicable]
   \item Discussion of sensible content if applicable, e.g., personally identifiable information or offensive content. [Not Applicable]
 \end{enumerate}

 \item If you used crowdsourcing or conducted research with human subjects, check if you include:
 \begin{enumerate}
   \item The full text of instructions given to participants and screenshots. [Not Applicable]
   \item Descriptions of potential participant risks, with links to Institutional Review Board (IRB) approvals if applicable. [Not Applicable]
   \item The estimated hourly wage paid to participants and the total amount spent on participant compensation. [Not Applicable]
 \end{enumerate}

 \end{enumerate}

\appendix
\onecolumn

\renewcommand{\thesection}{S\arabic{section}}
\renewcommand{\thesubsection}{S\arabic{section}.\arabic{subsection}}
\renewcommand{\theequation}{S\arabic{equation}}
\renewcommand{\thealgorithm}{S\arabic{algorithm}}

\newtheorem{manualcorrinner}{Corollary}
\newenvironment{manualcorr}[1]{%
  \renewcommand\themanualcorrinner{#1}%
  \manualcorrinner
}{\endmanualcorrinner}

\newtheorem{manualtheoreminner}{Theorem}
\newenvironment{manualtheorem}[1]{%
  \renewcommand\themanualtheoreminner{#1}%
  \manualtheoreminner
}{\endmanualtheoreminner}

\newtheorem{manualfigureinner}{Figure}
\newenvironment{manualfigure}[1]{%
  \renewcommand\themanualfigureinner{#1}%
  \manualfigureinner
}{\endmanualfigureinner}

\newtheorem{manualtableinner}{Table}
\newenvironment{manualtable}[1]{%
  \renewcommand\themanualtableinner{#1}%
  \manualtableinner
}{\endmanualtableinner}

In continuation to the main paper, we present theoretical proofs, more details on the experiments and some additional experimental results. Our key sections are listed as follows.
\begin{itemize}
    \item Theoretical proofs \textbf{S1}.
    \item Visualizing predictions of our conditional generator \textbf{S2.1}.
    \item More experimental details and additional results \textbf{S2.2}.
\end{itemize}
\section*{S1 THEORETICAL PROOFS}

\subsection*{S1.1 Proof of Lemma 1}\label{lemma1}
\textbf{Lemma1.}
Assuming $k$ is a normalized characteristic kernel, with probability at least $1-\delta$, we have:
$$\left|\int_{\calX\times\calY}\MMD^2\left(\pi_{Y|X}(\cdot|x), \delta_y\right)~ \textup{d}s_{X, Y}(x, y)-\frac{1}{m}\sum\limits_{i=1}^m\MMD^2\left(\pi_{Y|X}(\cdot|x_i), \delta_{y_i}\right) \right|
\le 2\sqrt{\frac{2}{m}\log\left(\frac{2}{\delta}\right)}.$$

\begin{proof}
Recall that MMD is nothing but the RKHS norm-induced distance between the corresponding kernel embeddings i.e., $\MMD(s,t)=\|\mu_k\left(s\right)-\mu_k\left(t\right)\|,$ where $\mu_k\left(s\right)\equiv \int \phi_k(x) ~\textup{d}s_X$, is the kernel mean embedding of $s$~\cite{MAL-060}, $\phi_k$ is the canonical feature map associated with the characteristic kernel $k$. Let $\calH_k$ denote the RKHS associated with the kernel $k$. 
Since our kernel is normalized we have that $\|\mu_k(b)\|\le1\ \forall\ b\in\calP(\calY)$. Hence, $0\leq \MMD^2\left(\pi_{Y|X}(\cdot|x), s_{Y|X}(\cdot|x)\right)=\|\mu_k\left(\pi_{Y|X}(\cdot|x)\right)-\mu_k\left(s_{Y|X}(\cdot|x)\right)\|^2\leq\|\mu_k\left(\pi_{Y|X}(\cdot|x)\right)+\mu_k\left(s_{Y|X}(\cdot|x)\right)\|^2\leq \left(\|\mu_k\left(\pi_{Y|X}(\cdot|x)\right)\|+\|\mu_k\left(s_{Y|X}(\cdot|x)\right)\|\right)^2 \le4$, where the second last step uses the triangle inequality. From Chernoff-Hoeffding bound, we have that: with probability at least $1-\delta$, $\left|\int_{\calX\times\calY}\MMD^2\left(\pi_{Y|X}(\cdot|x), \delta_{y}\right)\textup{d}s_{X,Y}(x, y)-\frac{1}{m}\sum_{i=1}^m\MMD^2\left(\pi_{Y|X}(\cdot|x_i), \delta_{y_i}\right)   \right|\le 2\sqrt{\frac{2}{m}\log\left(\frac{2}{\delta}\right)}$.
\end{proof}
\subsection*{S1.2 Proof of Theorem 1}\label{app:consistency}
We first restate Corollary (4) from the result of vector-contraction inequality for Rademacher in \cite{rad}, which we later use in our proof.
\begin{manualcorr}{(Restated from \cite{rad})}
Let $\calH$ denote a Hilbert space and let $f$ be a class of functions $f:\calX\mapsto \calH$, let $h_i:\calH\mapsto \R$ have Lipschitz norm $L$. Then
$$\E\sup_{f\in F} \sum_i \epsilon_i h_i(f(x_i))\leq \sqrt{2}L\sum_{i,k} \epsilon_{i,k} f_k(x_i),$$
where $\epsilon_{ik}$ is an independent doubly indexed Rademacher sequence, and $f_k(x_i)$ is
the $k$-th component of $f(x_i)$.
\end{manualcorr}
Our consistency theorem from the main paper is presented below, followed by its proof.
\paragraph{Proof of Theorem 1.}

\begin{proof}

From the definition of $\calU[\hat{\pi}_m]$ and $\calU[\pi^*]$, it follows that $0 \leq \calU[\hat{\pi}_m]-\calU[\pi^*]$.
\begin{align}
0\leq \calU[\hatpim]-\calU[\pi^*] & = \calU[\hatpim]-\hat{\calU}_m[\hatpim]+\hat{\calU}_m[\hatpim] - \hat{\calU}_m[\pi^*] + \hat{\calU}_m[\pi^*] - \calU[\pi^*] \nonumber \\
& \leq \calU[\hatpim]-\hat{\calU}_m[\hatpim]+ \hat{\calU}_m[\pi^*] - \calU[\pi^*] \ (\because \hatpim \textup{ is the solution of }\ref{eqn:empcot2}) \nonumber \\ 
& \leq \max_{\pi\in\Pi} \ (\calU[\pi]-\hat{\calU}_m[\pi])+ \hat{\calU}_m[\pi^*] - \calU[\pi^*] \label{thm1proof:1}
\end{align}
We now separately upper bound the two terms in \ref{thm1proof:1} : $(\hat{\calU}_m[\pi^*] - \calU[\pi^*])$ and $\max_{\pi\in\Pi} \ (\calU[\pi]-\hat{\calU}_m[\pi])$.
From Lemma \ref{lemma}, with probability at least $1-\delta$,
\begin{align} \label{thm1:1.1}
    \hat{\calU}_m[\pi^*] - \calU[\pi^*] \leq 2(\lambda_1+\lambda_2)\sqrt{\frac{2}{m}\log{\frac{2}{\delta}}}
\end{align}

We now turn to the second term. We show that $\max_{\pi\in\Pi} \calU[\pi]-\hat{\calU}_m[\pi]$ satisfies the bounded difference property. Let $Z_i$ denote the random variable $(X_i, Y_i)$. Let $Z=\{Z_1, \cdots, Z_i, \cdots, Z_m\}$ be a set of independent random variables. Consider another such set that differs only at the $i^{th}$ position : $Z'=\{Z_1, \cdots, Z_{i'}, \cdots, Z_m\}$. Let $\hat{\calU}_m[\pi]$ and $\hat{\calU}'_m[\pi]$ be the corresponding objectives in \ref{eqn:empcot2}.

\begin{align}
    &\left|\max_{\pi\in\Pi} \left(\calU[\pi]-\hat{\calU}_m[\pi]\right) - \max_{\pi\in\Pi} \left(\calU[\pi]-\hat{\calU}'_m[\pi]\right)\right| \nonumber \\
    &\leq \left|\max_{\pi\in\Pi} -\hat{\calU}_m[\pi] + \hat{\calU}'_m[\pi]\right| \nonumber\\
    &\leq \frac{\lambda_1}{m}\left|\max_{\pi\in\Pi} \MMD^2(\pi_{Y|X}(\cdot|x_i), \delta_{y_i})-\MMD^2(\pi_{Y|X}(\cdot|x'_i), \delta_{y'_i})\right|\nonumber\\&+\frac{\lambda_2}{m}\left|\max_{\pi\in\Pi} \MMD^2(\pi_{Y'|X}(\cdot|x_i), \delta_{y_i})-\MMD^2(\pi_{Y'|X}(\cdot|x'_i), \delta_{y'_i})\right| \nonumber 
    \ \textup{(Using triangle inequality)} \nonumber
    \\
    &\leq \frac{8(\lambda_1+\lambda_2)}{m},
\end{align}
where for the last step, we use that, with a normalized kernel, $\left(\textup{MMD}(\pi_{\bar{Y}}(\cdot|x_i), \delta_{y_i})+\textup{MMD}(\pi_{\bar{Y}}(\cdot|x_i'), \delta_{y_i'})\right)\leq 4$ and $ \left(\textup{MMD}(\pi_{\bar{Y}}(\cdot|x_i), \delta_{y_i})-\textup{MMD}(\pi_{\bar{Y}}(\cdot|x_i'), \delta_{y_i'})\right)\leq 2$ for $\bar{Y}\in\{Y, Y'\}$.

Using the above in McDiarmid's inequality,
\begin{align}\label{mcd}
    \max_{\pi\in\Pi} \calU[\pi]-\hat{\calU}_m[\pi] \leq \E\left[\max_{\pi\in\Pi} \calU[\pi]-\hat{\calU}_m[\pi]\right] + 4(\lambda_1+\lambda_2)\sqrt{\frac{2}{m}\log{\frac{1}{\delta}}}.
\end{align}
Let $Z_i\equiv(X_i, Y_i)\sim s_{X,Y}$ and $Z=\{Z_1, \cdots, Z_m\}$. Let $Z'_i\equiv(X'_i, Y'_i)\sim t_{X, Y}$ and $Z'=\{Z'_1, \cdots, Z'_m\}$. Let $(\epsilon_i)_{ i\in\{1, \cdots, m\} }$ be IID Rademacher random variables. We now follow the standard symmetrization trick and introduce the Rademacher random variables to get the following.
\begin{align}
\scriptstyle\E\left[\max_{\pi\in\Pi} \calU[\pi]-\hat{\calU}_m[\pi]\right] \leq 2\lambda_1\underbrace{\scriptstyle\frac{1}{m}\E_{Z, \epsilon}\left[\max_{\pi\in\Pi}\sum_{i=1}^m\epsilon_i\|\mu_k(\pi_{Y|X}(\cdot|X_i)) - \phi(Y_i)\|^2\right]}_{\calR_{m}(\Pi)}+\scriptstyle2\lambda_2\underbrace{\scriptstyle\scriptstyle\frac{1}{m}\E_{Z', \epsilon}\left[\max_{\pi\in\Pi}\sum_{i=1}^m\epsilon_i\|\mu_k(\pi_{Y'|X}(\cdot|X_i)) - \phi(Y_i)\|^2\right]}_{\calR'_{m}(\Pi)}. \label{Rm}
\end{align}Recall that $\mu_k(s)$ is the kernel mean embedding of the measure $s$. Hence, using \ref{thm1:1.1}, \ref{mcd} and \ref{Rm}, we prove that with probability at least $1-\delta$,
\begin{align}\label{eqn:apppacbound}
 \calU[\hatpim]-\calU[\pi^*]\leq 2\lambda_1\calR_{m}(\Pi) +2\lambda_2\calR'_{m}(\Pi)+ 6(\lambda_1+\lambda_2)\sqrt{\frac{2}{m}\log{\frac{3}{\delta}}}.
\end{align}

\textbf{Bounding Rademacher in the Special Case:} We now upper-bound $\calR_m(\Pi)$ for the special case where $\pi(\cdot|x)$ is implicitly defined using neural conditional generative models. More specifically, let $d$ be the dimensionality of $\calY$ and let $g_{\mathbf{w}}(x,N)\in\R^{2d}\sim\pi(\cdot|x)$, where $g_{\mathbf{w}}$ is a neural network function parameterized by $\mathbf{w}$, $N$ denotes the noise random variable. We make a mild assumption on the weights of the neural network to be bounded. The first $d$ outputs, denoted by $g_{\mathbf{w},1}(x,N)$ will be distributed as $\pi_{Y|X}(\cdot|x)$ and the last $d$ outputs, denoted by $g_{\mathbf{w},2}(x,N)$ will be distributed as $\pi_{Y'|X}(\cdot|x)$.
Let $\zeta_i(\pi_{Y|X})\equiv \|\mu_k(\pi_{Y|X}(\cdot|x_i)) - \phi(y_i)\|^2$. We now compute the Lipschitz constant for $\zeta_i$, used in our bound next. 
\begin{align}
    \zeta_i(\pi_{Y|X})-\zeta_i(\pi_{Y|X}') 
    & \leq 4\left( \|\mu_k\left(\pi_{Y|X}(\cdot|x_i)\right)-\phi(y_i)\| - \|\mu_k\left(\pi'_{Y|X}(\cdot|x_i)\right)-\phi(y_i) \| \right) \ \textup{(With a normalized kernel)}\nonumber\\
    &\leq  4\|\mu_k(\pi_{Y|X}(\cdot|x_i))-\mu_k(\pi'_{Y|X}(\cdot|x_i))\|\ \textup{(Using triangle inequality)} \nonumber \\
    & =4\|\E\left[\phi(g_{\mathbf{w},1}(x_i,N))\right]-\E\left[\phi(g_{\mathbf{w}',1}(x_i,N))\right]\|\nonumber\\
   & \le4\E\left[\|\phi(g_{\mathbf{w},1}(x_i,N))-\phi(g_{\mathbf{w}',1}(x_i,N))\|\right]\ \because \textup{ (Jensen's inequality)}\nonumber\\
   & \le4\E\left[\|g_{\mathbf{w},1}(x_i,N)-g_{\mathbf{w}',1}(x_i,N)\|\right]\ \because \textup{ (non-expansive kernel)}\nonumber\\
   & \le4\left[\|g_{\mathbf{w},1}(x_i,n_{i,1})-g_{\mathbf{w}',1}(x_i,n_{i,1})\|\right]\ \because n_{i,j}\equiv\arg\max_{n}\left[\|g_{\mathbf{w},j}(x_i,n)-g_{\mathbf{w}',j}(x_i,n)\|\right].\nonumber\\
\end{align}

We next use a vector-contraction inequality for Rademacher given in Corollary (4) from \cite{rad}. This gives $\calR_m(\Pi)\le\frac{4\sqrt{2}}{m}\E_{Z,\epsilon}\max\limits_\mathbf{w}\sum_{i=1}^m\sum_{j=1}^{d}r_{ij}g^j_{\mathbf{w},1}(x_i,n_{i,1})$ and $\calR'_m(\Pi)\le\frac{4\sqrt{2}}{m}\E_{Z',\epsilon}\max\limits_\mathbf{w}\sum_{i=1}^m\sum_{j=1}^{d}r_{ij}g^j_{\mathbf{w},2}(x_i,n_{i,2})$. Here, $g^j_{\mathbf{w},1}, g^j_{\mathbf{w},2}$ denote the $j^{th}$ output in the first and the second blocks; $r_{ij}$ denotes an independent doubly indexed Rademacher variable. Thus, we have upper bounded the complexity of $\Pi$ in terms of that of the neural networks.

Now, applying standard bounds (e.g. refer to $\S 5$ in \cite{neyshabur2017implicit}) on Rademacher complexity of neural networks, we obtain $\calR_m(\Pi)\le O(1/\sqrt{m})$ and $\calR'_m(\Pi)\le O(1/\sqrt{m})$. If $\lambda_1,\lambda_2$ are chosen to be $O(m^{1/4})$, then from (\ref{eqn:apppacbound}), we have: $\calU[\hatpim]-\calU[\pi^*]\leq O(1/m^{1/4})$. When $m\rightarrow\infty$, this shows that $\hatpim$ is also an optimal solution of (\ref{eqn:empcot2}), in which case it is also an optimal solution of the original COT problem (when restricted to $\Pi$) because $\lambda\rightarrow\infty$ too.
\end{proof}

\section*{S2 MORE ON EXPERIMENTS}\label{app:exp}
This section contains more experimental details along with some additional results.
\subsection*{S2.1 Visualizing Predictions of the Conditional Generator}
We visualize the predictions learnt by the implicit conditional generator trained with the COT loss $\ref{eqn:impcot}$ and the alternate formulation \ref{alt-form} described below. The COT formulation \ref{eqn:regcot1} employs a clever choice of MMD regularization over the conditionals, which is then computed using the samples from the joints \ref{eqn:regcot2}. One may think of alternatively employing an MMD regularization over joints as follows.
\begin{align}\label{alt-form}
\min_{\pi_{Y,Y'|X}:\calX\mapsto\calP(\calY\times\calY)}\int_{\calX}\int_{\calY\times\calY} c\ \textup{d}\pi_{Y,Y'|X}(\cdot, \cdot|x)\textup{d}a(x) + &\lambda_1\MMD^2\left(\pi_{Y|X}(\cdot|x)s(x), s(x, y)\right)\nonumber\\
+&\lambda_2\MMD^2\left(\pi_{Y'|X}(\cdot|x)t(y), t(x, y)\right).
\end{align}
We argue that this choice is sub-optimal. We first note that as we only have samples from the joints and not the marginal distributions ($s_X$ and $t_X$), matching conditionals through the above formulation is not straightforward. Computing the above formulation also incurs more memory because for computing the Gram matrix over $(x, y)$, we need to keep Gram matrices over the samples of $x$, $y$ separately. Further, in this case, each of the Gram matrices is larger than the ones needed with the proposed formulation \ref{eqn:regcot2}. We compared the performances of the two formulations in a toy regression case and found the proposed COT formulation better. 

\begin{figure}[t]
\centering
\includegraphics[width=\columnwidth]{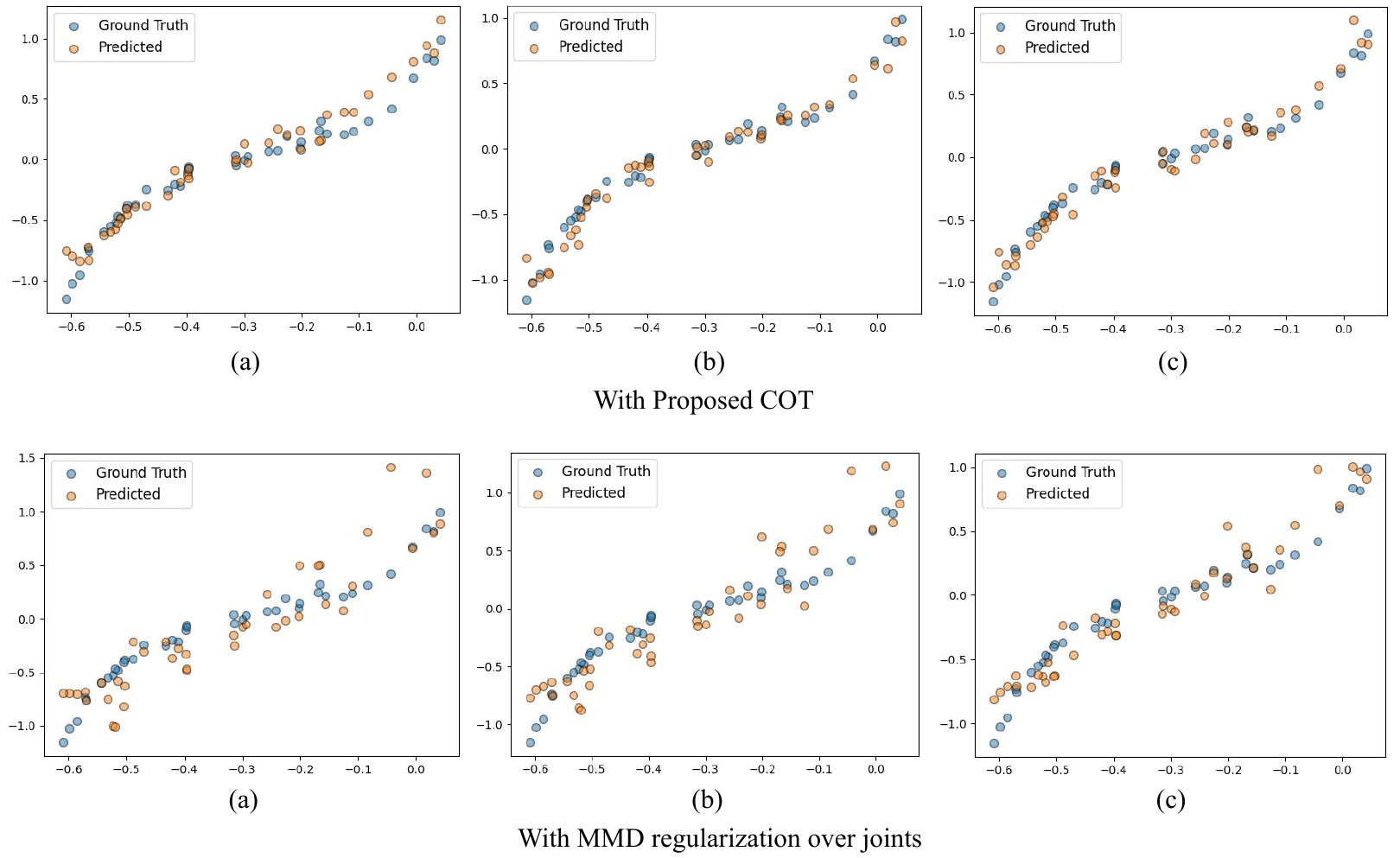}
\caption{Predictions of the implicit conditional generator trained with the COT loss \ref{eqn:impcot} and the alternate formulation \ref{alt-form} (with MMD regularization over joints). The plots show the effect of different $\sigma^2$ hyperparameters used in the RBF kernel as 1, 10 and 100, respectively. We quantitatively evaluate the methods using Explained Variance (between $-\infty$ and 1; higher is better). With the proposed COT loss, the explained variance scores are 0.94, 0.94 and 0.95, respectively. With the alternate formulation \ref{alt-form}, the explained variance scores are 0.63, 0.73 and 0.85. This shows the superiority of the proposed COT formulation \ref{eqn:impcot}.}
\label{supp:reg}
\end{figure}

The training algorithm for learning with the proposed COT loss is presented in Algorithm~\ref{algo-imp}. The per-epoch computational complexity is $O(m^2)$, where $m$. We fix $\lambda$ to 500, noise dimension to 10. We use Adam optimizer with a learning rate of $5e-3$ and train for 1000 epochs. We use squared Euclidean distance and RBF kernel. Figure \ref{supp:reg} shows we obtain a good fit for $\sigma^2=10, 100$.
\begin{algorithm}[t]
        \caption{Algorithm for learning with implicit models for a simple regression case.}
        \label{algo-imp}
\begin{algorithmic}[1]
        \Require Implicit neural networks $\pi_{\theta}$: $\mathcal{ X} \mapsto \mathcal{ Y}$ and $\pi_{\psi}$: $\mathcal{ X}, \mathcal{ Y} \mapsto \mathcal{ Y} $, training samples $(x_i, y_i)|_{i=1}^m$, noise distribution $\eta$, cost function $c: \mathcal{ Y} \times \mathcal{ Y}\mapsto \R^+$, kernel, $\lambda$.
 
        \While{not converged or max epochs not reached}
        \State Sample $z_i\sim \eta \ \forall i \in [m]$.
        \State $y_i(x_i; \theta)=\pi_\theta(\cdot|x_i, z_i) \forall i\in [m]$.  
        \State Sample $z'_{i}\sim \eta \  \forall i\in [m]$.
        \State $y_i (x_i; \theta, \psi) = \pi_\psi(\cdot|y_i (x_i; \theta), x_i, z'_{i}) ;\  \forall i\in [m]$.
        \State Compute the COT loss (Simplified case of Equation~\ref{eqn:impcot})
        \begin{align*} 
        \min_{\theta, \psi}&\frac{1}{m}\sum_{i=1}^m c\left(y_{i}\left(x_i;\theta\right), y_i\left(x_i;\theta,\psi\right)\right) +\lambda\frac{1}{m}\sum_{i=1}^m\textup{MMD}^2\left(\frac{1}{m}\sum_{j=1}^m\delta_{y_{j}\left(x_i;\theta,\psi\right) }, \delta_{y_i}\right).
        \end{align*}
        \State Update $\theta, \psi$ using gradient descent. 
         \EndWhile   
\end{algorithmic}
\end{algorithm}

In Table \ref{table:time-imp}, we also show the per-epoch computation time taken (on an RTX 4090 GPU) by the COT loss as a function of the size of the minibatch, which shows the computational efficiency of the COT loss. On the other hand, the computation time for the alternate formulation discussed in \ref{alt-form} (with MMD regularization over joints) is 0.245 $\pm$ 0.0012 s with minibatch-size 16 and resulted in the out-of-memory error for higher batchsizes.

\begin{minipage}{\textwidth}

  \begin{minipage}[b]{0.3\textwidth}
    \centering
    \includegraphics[scale=0.35]{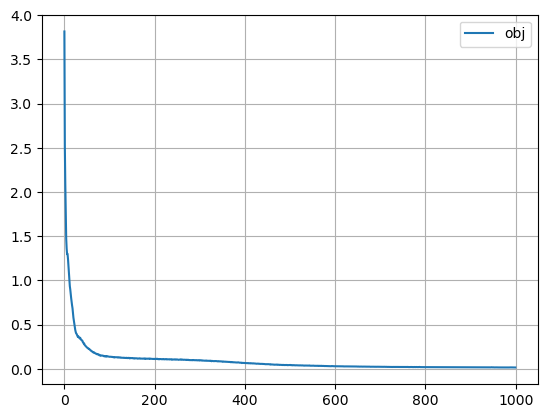}
    \captionof{figure}{The objective over the training epochs curve.}
  \end{minipage}
\hfill
\begin{minipage}[b]{0.6\textwidth}
\centering
      \begin{tabular}{cc}
        \toprule
          $B$ & Time (s) \\  
        \midrule
         16 & 0.228$\pm$0.0003\\
         64 & 0.228$\pm$0.0004\\
         512 & 0.229$\pm$0.0008\\
         1024 & 0.231$\pm$0.0016 \\
        \bottomrule 
        \end{tabular}
      \captionof{table}{Time (in s) for COT loss \ref{eqn:impcot} computation shown for increasing minibatch size ($B$). The computation time reported is based on 3 independent runs on the toy regression dataset.}\label{table:time-imp}
    \end{minipage}
\end{minipage}

\subsection*{S2.2 More Experimental Details}
We provide more details for the experiments shown in $\S~\ref{sec:exp}$ of the main paper, along with some additional results.
\paragraph{Verifying the Correctness of Estimator}
We use Adam optimizer and jointly optimize $\pi_\theta$ and $\pi_\psi$. We choose $\lambda$ from the set \{1, 200, 500, 800, 1000\} and $\sigma^2$ used in the RBF kernel from the set \{1e-2, 1e-1, 1, 10\}. We found $\lambda$ as 1000 and $\sigma^2$ as 1 to perform the best.

\begin{figure}
    \centering
    \includegraphics[scale=0.3]{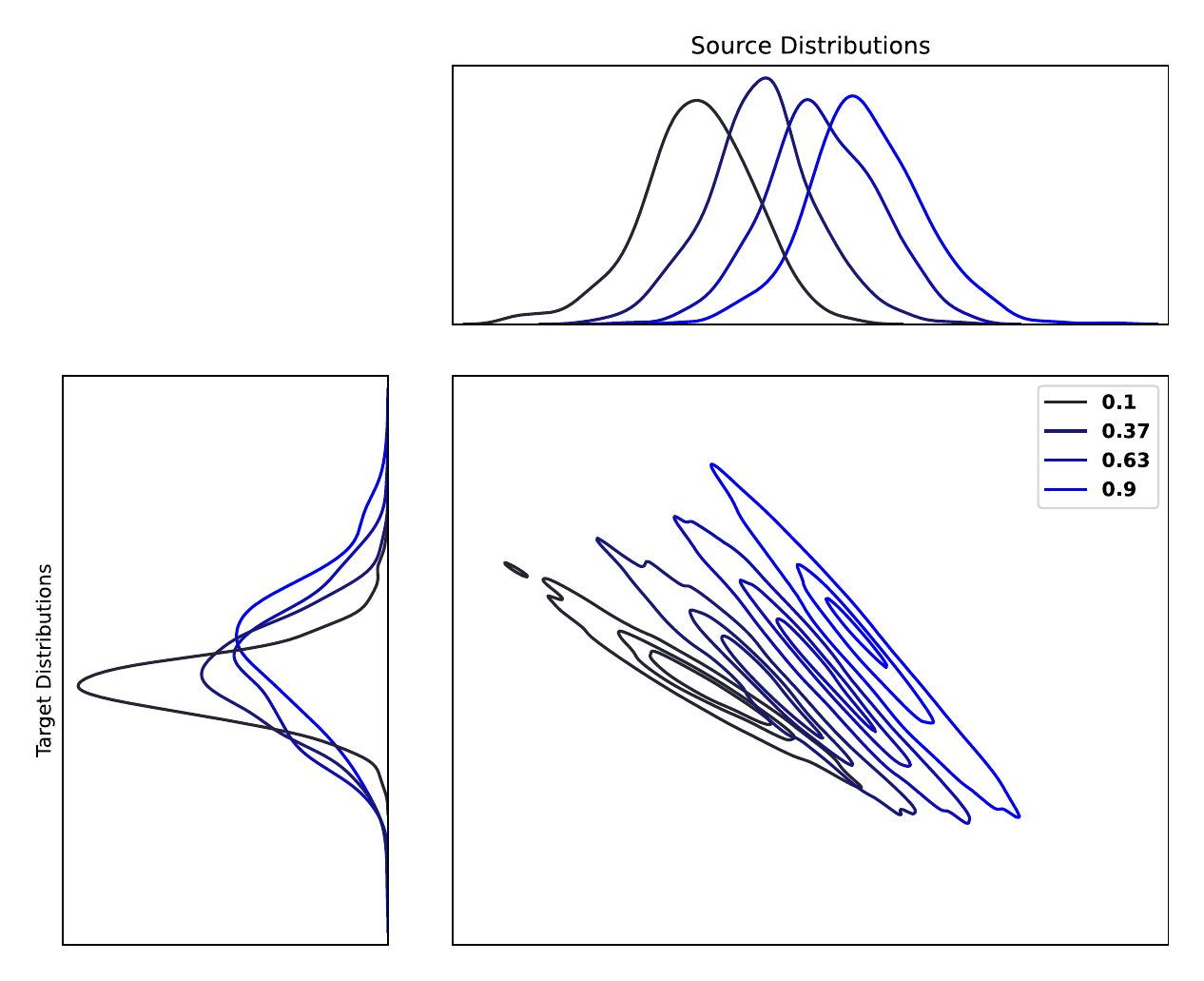}
    \caption{The OT plans computed with the COT formulation \ref{eqn:impcot} for the case of source and target as the conditional Gaussian distributions. For each value of the conditioned variable, we show the corresponding source, target and the obtained OT plan in a given color.}
    \label{fig:plan}
\end{figure}

In Figure \ref{fig:plan}, we also show the OT plans. We draw 500 samples from the implicit maps learnt with the COT loss \ref{eqn:impcot} and use kernel density estimation (KDE) to plot the distributions.

\paragraph{Cell Population Dynamics}
Dataset: We use the preprocessed dataset provided by \cite{cellot}. The dataset is publicly available for download using the following link \begin{verbatim}https://polybox.ethz.ch/index.php/s/RAykIMfDl0qCJaM.\end{verbatim} From this dataset, we extracted unperturbed cells and cells treated with \texttt{Givinostat}. This led to a total of 17565 control cells and a total of 3541 cells treated with \texttt{Givinostat}.  We take the same data splits as in \cite{cellot}.

More on evaluation: Following \cite{Cuturi22}, we use \texttt{scanpy}'s \cite{Wolf2018} \texttt{rank\_genes\_groups} function for ranking and obtaining 50 marker genes for the drug, in this case \texttt{Givinostat}. The perturbed cells are grouped by drug, and the ranking is computed by keeping the unperturbed (i.e. control) cells as reference. We fix the architecture of our implicit model ($\psi$) as a 5-layer MLP and train it for 1000 epochs. Similar to \cite{Cuturi22}, we train on the 50-dimensional representation after applying PCA on the 1000-dimensional original representation. It is worth noting that training our MLP models is much stabler than the Partial Input Convex Neural Networks (PICNN) used in \cite{Cuturi22}, which needs carefully chosen initialization schemes. Following the evaluation scheme in \cite{Cuturi22}, we get back to the original 1000 dimensions, and then 50 marker genes are computed for the evaluation metrics.

Following the in-sample experiment done in \cite{Cuturi22}, we tune our hyperparameters on the training data split. Based on the scale of terms in the COT objective, we chose $\lambda$ from the set $\{400, 2000, 10000\}$ and found $\lambda=400$ to be the optimal choice. For the IMQ kernel, we chose the hyperparameter from the set $\{1, 10, 50, 100\}$ and found 100 to be the optimal choice. Since we model the transport plan and not the transport map, the following procedure is followed for inference. We generate one sample corresponding to each pair of (source sample, condition) through our implicit model, and measure the required metrics on the generated distributions. This procedure is repeated $n=50$ times, and the average metric is reported.

Following \cite{Cuturi22}, we quantitatively evaluate our performance using the MMD distance and the $l_2$ distance between the perturbation signatures, $l_2$(PS) metric. Let $\mu$ be the set of observed unperturbed cell population, $\nu$ be the set of the observed perturbed cell population (of size $m_1$), and $\nu'$ be the set of predicted perturbed state of population $\mu$ (of size $m_2$). The perturbation signature $\textup{PS}(\nu, \mu)$ is then defined as 
$\frac{1}{m_1}\sum_{y_i\in\nu}y_i-\frac{1}{m_2}\sum_{y_i\in\mu}y_i'$. The $l_2$(PS) metric is the $l_2$ distance between $\textup{PS}(\nu, \mu)$ and $\textup{PS}(\hat{\nu}, \mu)$. Following \cite{Cuturi22}, we report MMD ($\S~\ref{sec:pre}$) with RBF kernel averaged over the kernel widths: \{2, 1, 0.5, 0.1, 0.01, 0.005\}.

\begin{table}[t]
  \caption{Insample setting: $l_2$ (PS) distances (lower is better) between predicted and ground truth distributions where the marker genes are computed at a per-dose level.}
  \label{sample-table-l2-perdose}
  \centering
  \footnotesize{
  \begin{tabular}{lccc}
    \toprule
        Dosage   & CellOT & CondOT & Proposed \\  
        \midrule
        $10nM$ & 0.7164 & 0.4718 & \textbf{0.3682} \\
        $100nM$ & 0.5198 & 0.3267 & \textbf{0.3051}\\
        $1000nM$ & 0.7075 & 0.6982 & \textbf{0.3917} \\
        $10000nM$ & 4.8131 & 0.3457 & \textbf{0.2488}  \\
        \textbf{Average} & 1.6892 & 0.4606 &  \textbf{0.3284}\\
        \bottomrule 
  \end{tabular}}
\end{table}

\begin{table}[t]
  \caption{Insample setting: MMD distances (lower is better) between predicted and ground truth distributions where the marker genes are computed at a per-dose level.}
  \label{sample-table-mmd-perdose}
  \centering
  \footnotesize{
  \begin{tabular}{lccc}
    \toprule
        Dosage   & CellOT & CondOT & Proposed \\  
        \midrule
        $10nM$ & 0.0089 & 0.0064 & \textbf{0.00549} \\
        $100nM$ & 0.0069 & 0.0054 & \textbf{0.00494}\\
        $1000nM$ & 0.0117 & 0.01038 & \textbf{0.00586} \\
        $10000nM$ & 0.16940 & 0.01051 & \textbf{0.01011}  \\
        \textbf{Average} & 0.04922 & 0.00817 &  \textbf{0.00660}\\
        \bottomrule 
  \end{tabular}}
\end{table}

\begin{figure}
    \centering
    \includegraphics[width=0.3\textwidth]{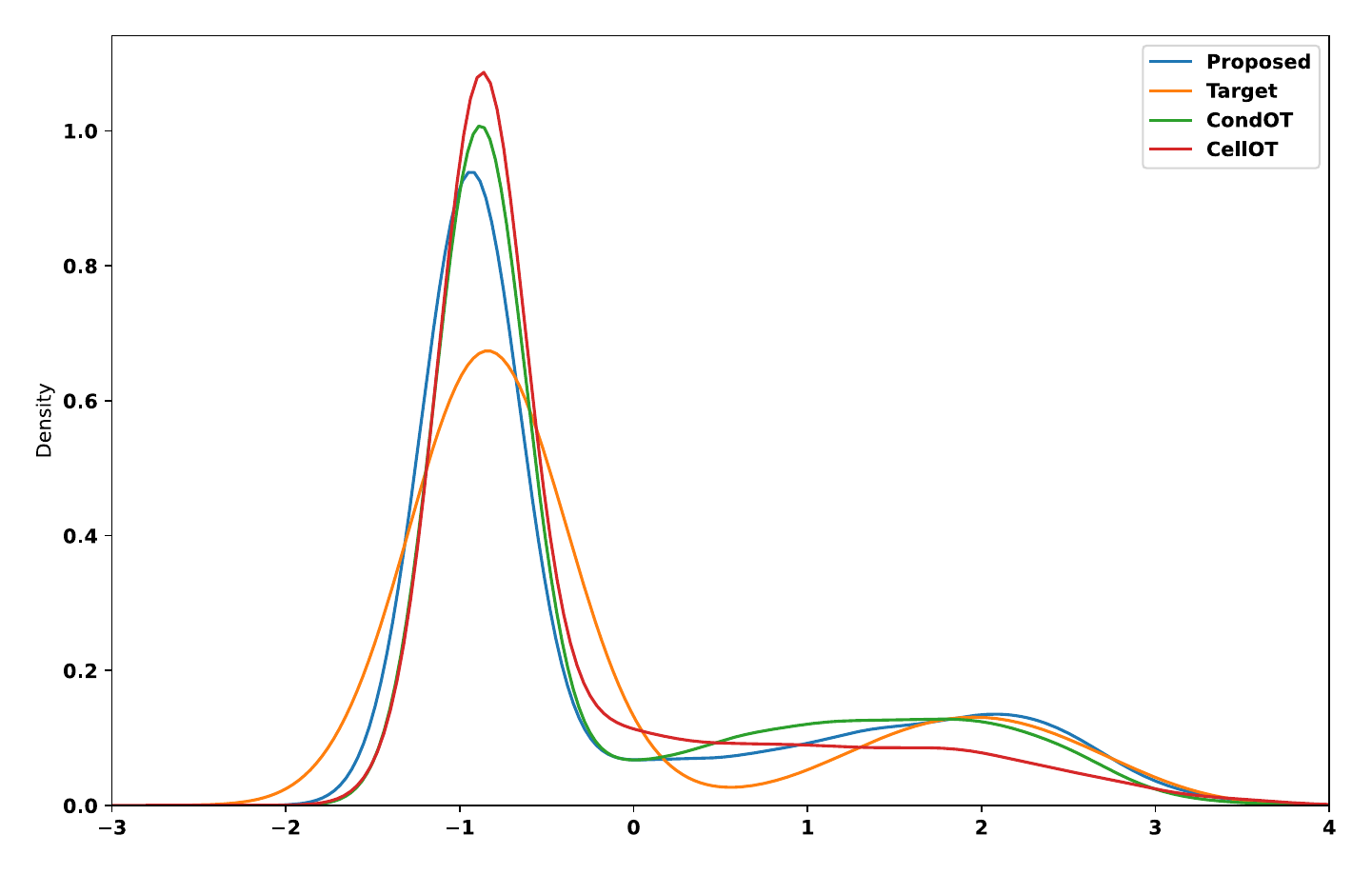}
    \includegraphics[width=0.3\textwidth]{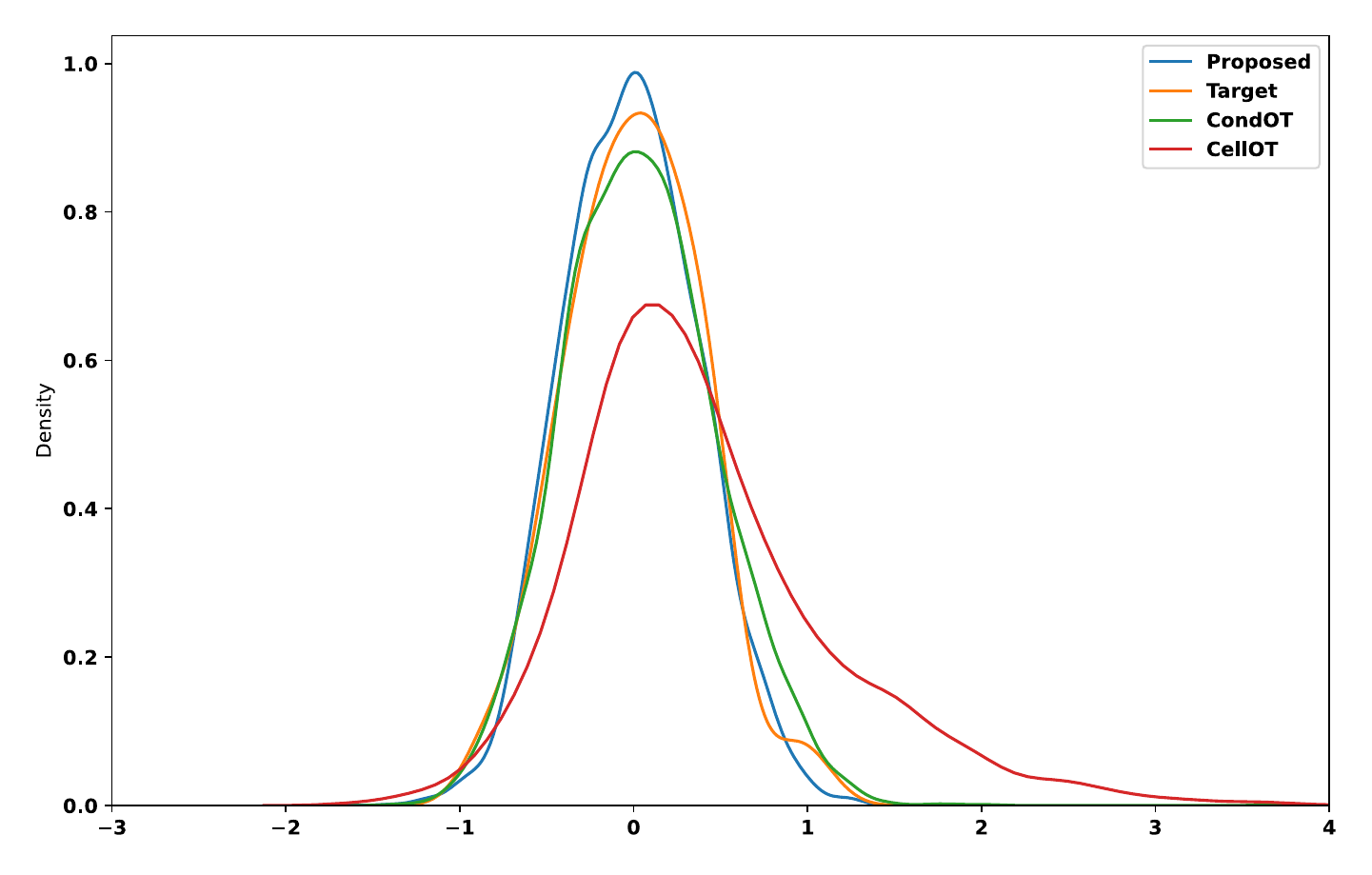}
    \includegraphics[width=0.3\textwidth]{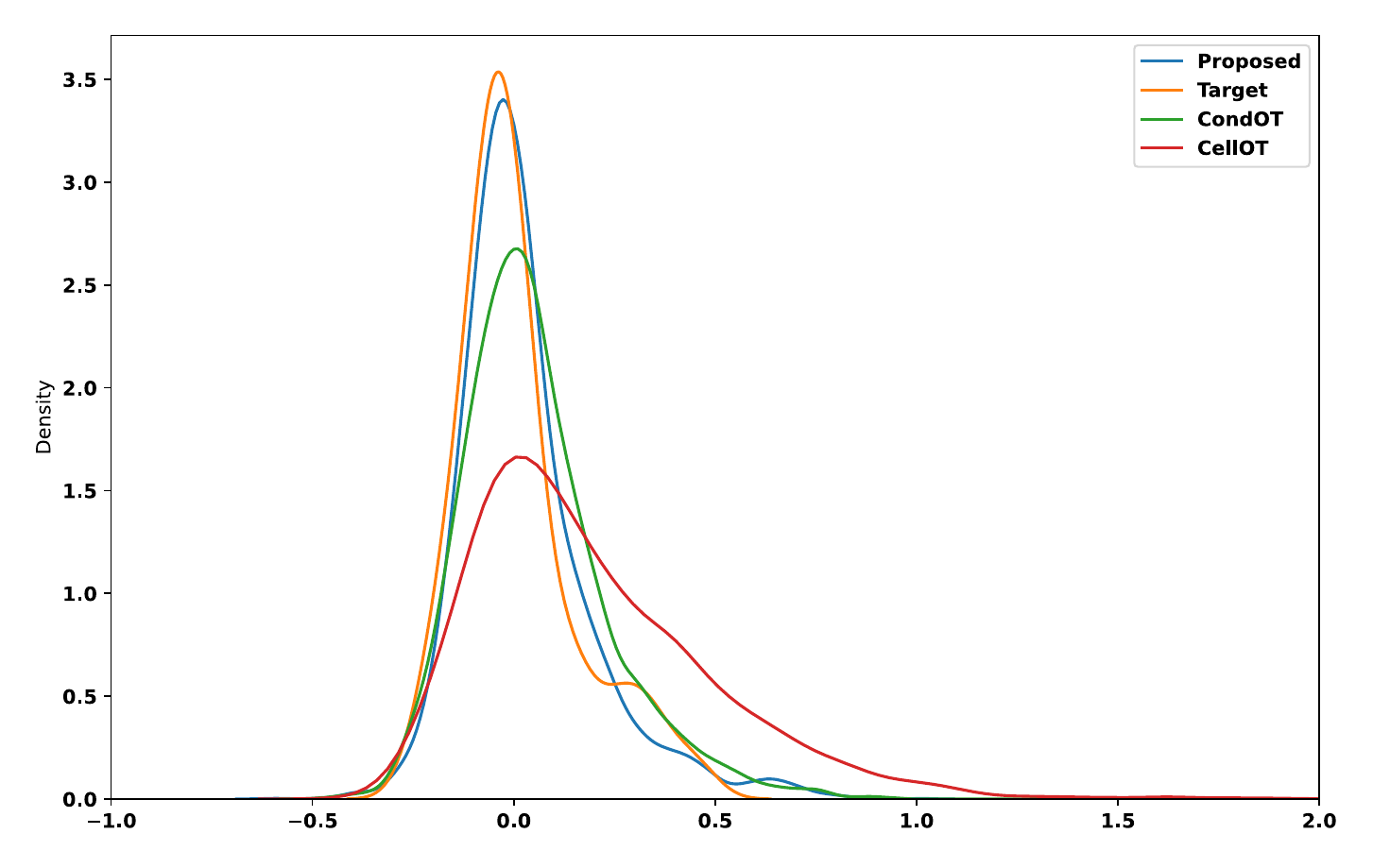}
    \caption{Marginals for selected genes `ENSG00000165092.12', `ENSG00000175175.5', `ENSG00000173727.12',  where the dosage is 100nM, in the insample setting.}
    \label{fig:marginal}
\end{figure}

\begin{figure}
    \centering
    \includegraphics[width=0.3\textwidth]{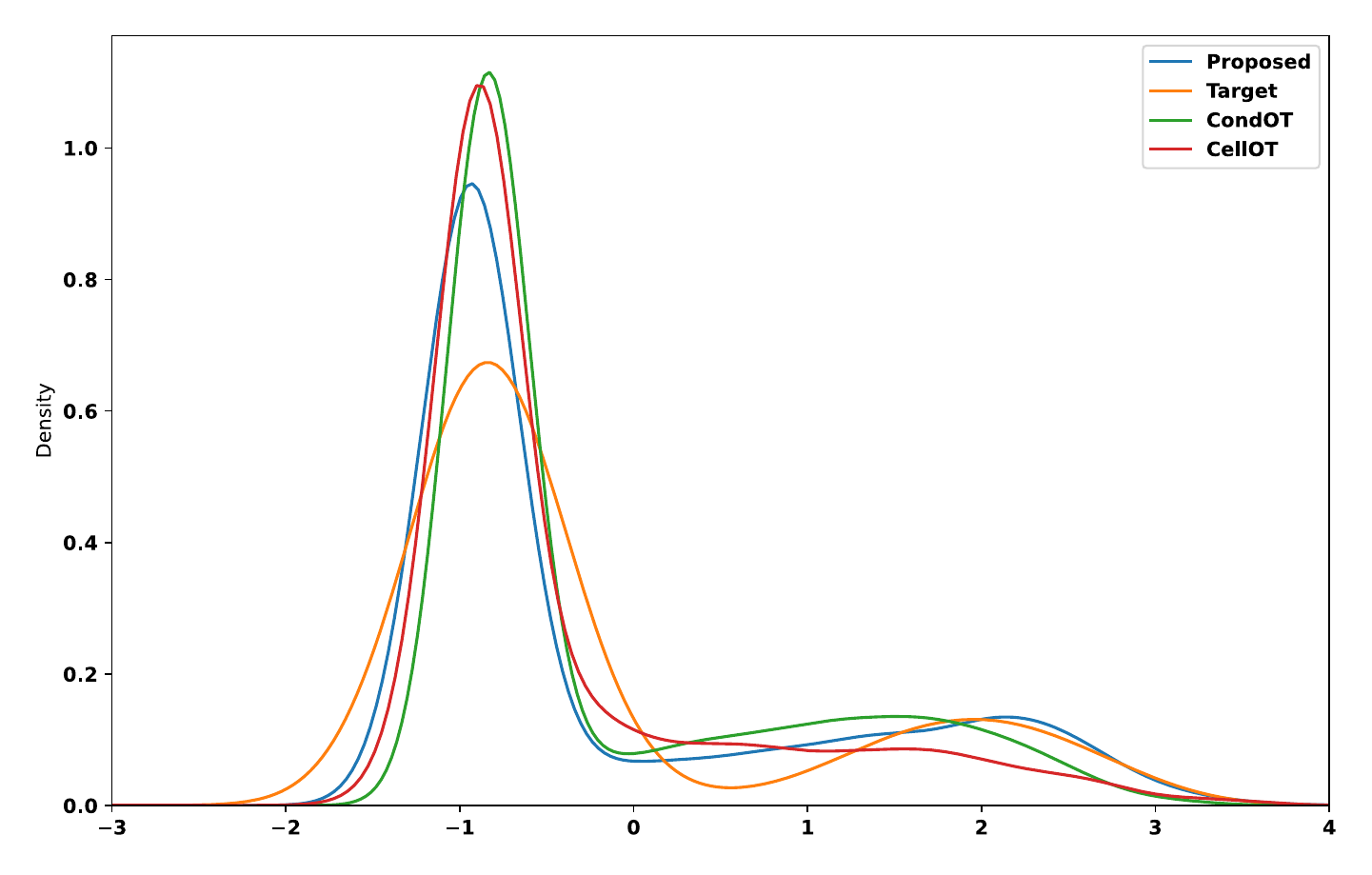}
    \includegraphics[width=0.3\textwidth]{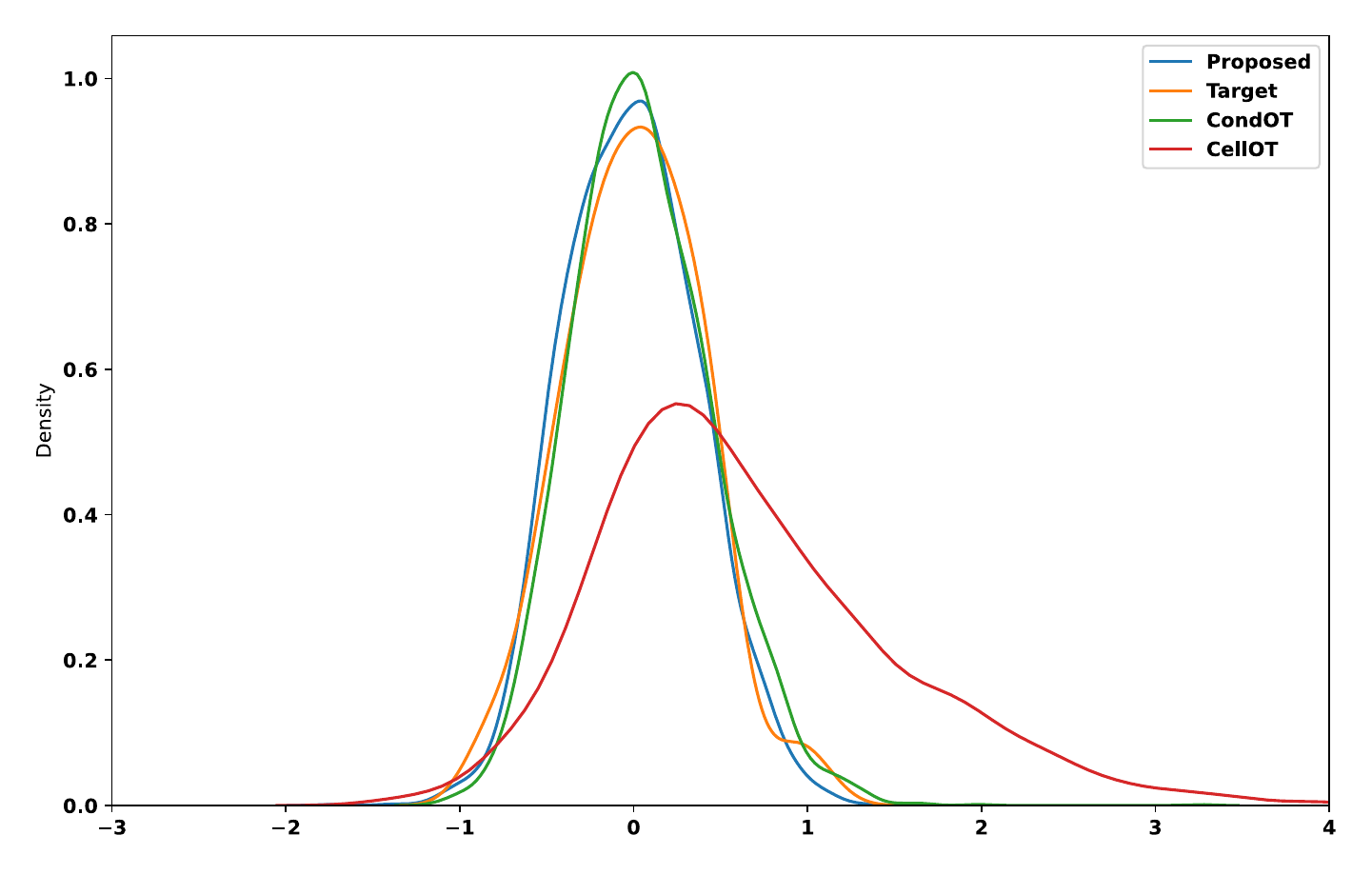}
    \includegraphics[width=0.3\textwidth]{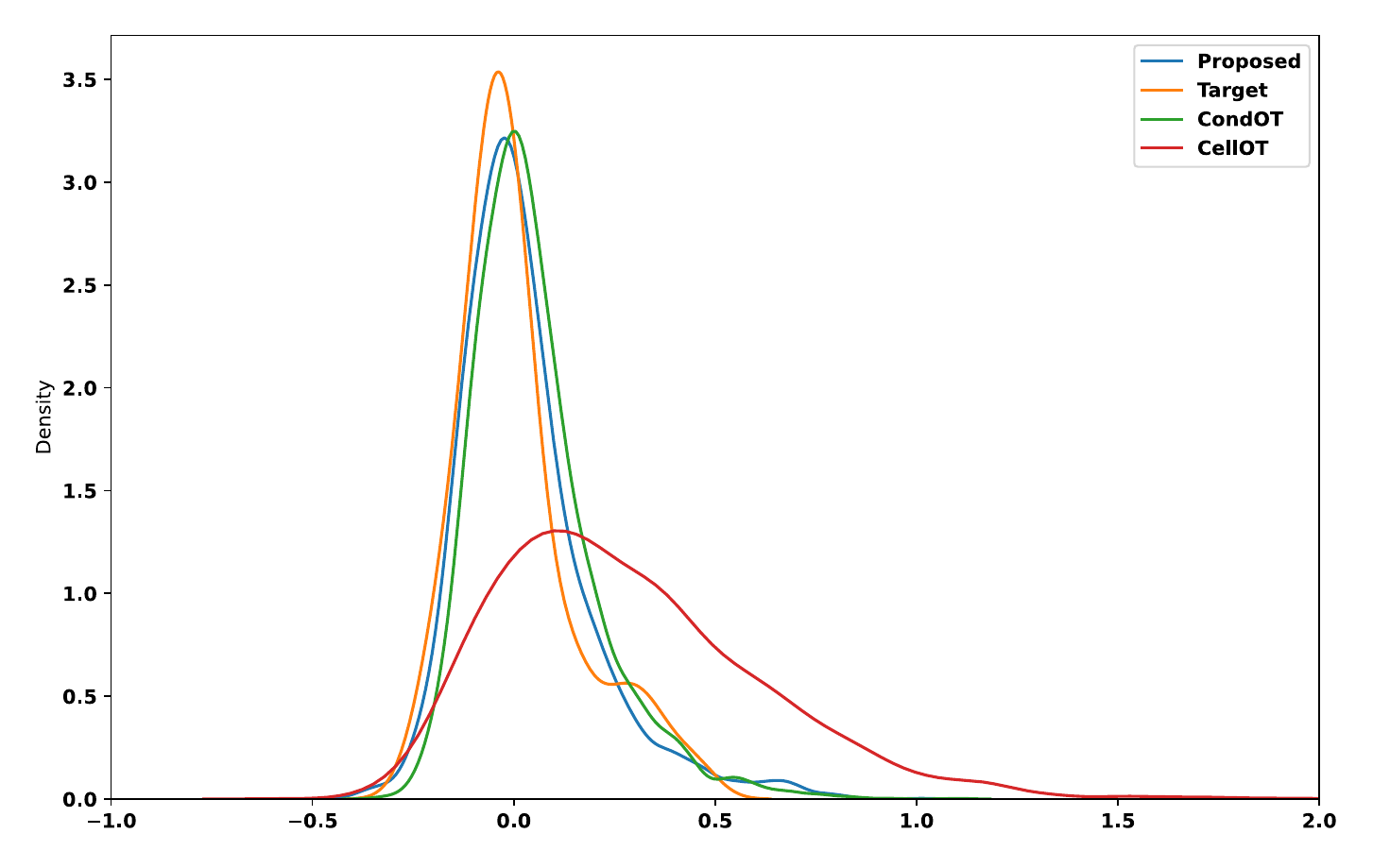}
    \caption{Marginals for selected genes `ENSG00000165092.12', `ENSG00000175175.5', `ENSG00000173727.12',  where the dosage is 100nM, in the outsample setting.}
    \label{fig:marginal-o}
\end{figure}

Additional Results: In addition to the results reported in Tables \ref{sample-table-l2} and \ref{sample-table-mmd} where the marker genes are computed on a per-drug level, in Tables \ref{sample-table-l2-perdose} and \ref{sample-table-mmd-perdose}, we show results where marker genes are computed on a per-dosage level.
Further, we present results for the out-of-sample setting, i.e., the dosage levels we predict are not seen during training. In Tables \ref{oos-l2} and \ref{oos-mmd}, we show the results when marker genes are computed on a per-drug level and in Tables \ref{oos-l2-dose} and \ref{oos-mmd-dose}. we show the results when marker genes are computed on a per-dose level. In Figures \ref{fig:marginal} and \ref{fig:marginal-o}, we also show how closely the marginals of the proposed conditional optimal transport plan match the target distribution. The plots for COT correspond to the generated distribution having the median value for the metrics among all the (n=50) generated distributions. 
\begin{table}[t]
  \caption{Out-of-sample setting: $l_2$ (PS) distances (lower is better) between predicted and ground truth distributions where the marker genes are computed at a per-drug level.}
  \label{oos-l2}
  \centering
  \footnotesize{
  \begin{tabular}{lccc}
    \toprule
        Dosage   & CellOT & CondOT & Proposed \\  
        \midrule
        $10nM$ & 2.0889 & 0.3789 & \textbf{0.3376} \\
        $100nM$ & 2.0024 & 0.2169 & \textbf{0.1914}\\
        $1000nM$ & 1.2596 & \textbf{0.9928} & 1.002 \\
        $10000nM$ & \textbf{5.9701} & 34.9016 & 8.2417  \\
        \bottomrule 
  \end{tabular}}
\end{table}

\begin{table}[t]
  \caption{Out-of-sample setting: MMD distances (lower is better) between predicted and ground truth distributions where the marker genes are computed at a per-drug level.}
  \label{oos-mmd}
  \centering
  \footnotesize{
  \begin{tabular}{lccc}
    \toprule
        Dosage   & CellOT & CondOT & Proposed \\  
        \midrule
        $10nM$ & 0.0369 & \textbf{0.0065} & 0.0071 \\
        $100nM$ & 0.0342 & \textbf{0.0061}  & 0.0070\\
        $1000nM$ & 0.0215 & 0.0178 & \textbf{0.0151} \\
        $10000nM$ & \textbf{0.2304} & 0.3917 & 0.3591  \\
        \bottomrule 
  \end{tabular}}
\end{table}

\begin{table}
  \caption{Out-of-sample setting: $l_2$ (PS) distances (lower is better) between predicted and ground truth distributions where the marker genes are computed at a per-dose level.}
  \label{oos-l2-dose}
  \centering
  \footnotesize{
  \begin{tabular}{lccc}
    \toprule
        Dosage   & CellOT & CondOT & Proposed \\  
        \midrule
        $10nM$ & 1.2130 & 0.4718 & \textbf{0.3950} \\
        $100nM$ & 0.8561 & 0.2846 & \textbf{0.2522}\\
        $1000nM$ & \textbf{0.9707} & 0.9954 & 1.0775 \\
        $10000nM$ & \textbf{5.8737} & 33.5211 & 7.1487  \\
        \bottomrule 
  \end{tabular}}
\end{table}

\begin{table}
  \caption{Out-of-sample setting: MMD distances (lower is better) between predicted and ground truth distributions where the marker genes are computed at a per-dose level.}
  \label{oos-mmd-dose}
  \centering
  \footnotesize{
  \begin{tabular}{lccc}
    \toprule
        Dosage   & CellOT & CondOT & Proposed \\  
        \midrule
        $10nM$ & 0.01648 & 0.00641 & \textbf{0.00638} \\
        $100nM$ & 0.01133 & 0.006325  & \textbf{0.00571}\\
        $1000nM$ & 0.01607 & 0.01496 & \textbf{0.01462} \\
        $10000nM$ & \textbf{0.24234} & 0.41845 & 0.34246  \\
        \bottomrule 
  \end{tabular}}
\end{table}

\paragraph{Classification} We consider the task of multi-class classification and experiment on three benchmark datasets MNIST \cite{mnist}, CIFAR-10 \cite{krizhevsky2009learning} and Animals with Attribute (AWA) \cite{5206594}. Following the popular approaches of minibatch OT \cite{fatras2019learnwass, fatras2021jumbot}, we perform a minibatch training. We use the implementation of \cite{Frogner15} open-sourced by \cite{drpratikrot4c}. We maintain the same experimental setup used in \cite{drpratikrot4c}. The classifier is a single-layer neural network with Softmax activation trained for 200 epochs. We use the cost function, $c$, between labels as the squared $l_2$ distance between the fastText embeddings \cite{bojanowski-etal-2017-enriching} of the labels. The kernel function used in COT is $k(x, y) = 1/(\sigma^2+c(x, y))^{0.5}$. For MNIST and CIFAR-10, we use the standard splits for training and testing and choose a random subset of size 10,000 from the training set for validation. For AWA, we use the train and test splits provided by \cite{drpratikrot4c} and randomly take 30\% of the training data for validation.

Following \cite{drpratikrot4c}, we compare all methods using the Area Under Curve (AUC) score of the classifier on the test data after finding the best hyperparameters on the validation data. Based on the validation phase, the best Sinkhorn regularization hyperparameter in $\epsilon$-OT \cite{Frogner15} is 0.2. For COT, we choose the hyperparameters ($\lambda, \sigma^2$) based on the validation set: for MNIST (0.1, 0.1), for CIFAR-10 (0.1, 0.1) and for AWA (10, 0.1).

In Table \ref{table:time-exp}, we also show the per-epoch computation time taken (on an RTX 4090 GPU) by the COT loss as a function of the size of the minibatch, which shows the computational efficiency of the COT loss.

\begin{table}[t]
  \caption{Time (in s) for COT loss \ref{eqn:expexpcot} computation shown for increasing minibatch size. The computation time reported is based on 3 independent runs on the CIFAR-10 dataset.}
  \label{table:time-exp}
  \centering
  \begin{tabular}{ccccc}
    \toprule
           & 16 & 64 & 512 & 1024 \\  
        \midrule
        Time (s) & 0.229$\pm$0.0013 & 0.229$\pm$0.0006 & 0.227$\pm$0.0004 & 0.225$\pm$0.0021 \\
        \bottomrule 
  \end{tabular}
\end{table}

\paragraph{Prompt Learning}
Let $\mathbf{F}=\{\mathbf{f}_m|_{m=1}^M\}$ denote the set of visual features for a given image and $\mathbf{G}_r=\{\mathbf{g}_n|_{n=1}^N\}$ denote the set of textual prompt features for class $r$. PLOT \cite{chen2023plot} learns the prompt features by performing an alternate optimization where the inner optimization solves an OT problem between the empirical measure over image features (49) and that over the prompt features (4). We denote the OT distance between the visual features of image $\mathbf{x}$ and the textual prompt features of class $r$ by $d_{OT}(\mathbf{x}, r)$. Then the probability of assigning the image $\mathbf{x}$ to class $r$ is computed as
$
    p(y=r|\mathbf{x}) = \frac{\exp{\left((1-d_{OT}(\mathbf{x}, r)/\tau)\right)}}{\sum_{r=1}^T\exp{\left((1-d_{OT}(\mathbf{x}, r)/\tau)\right)}},
$ where $T$ denotes the total no. of classes and $\tau$ is the temperature of softmax. These prediction probabilities are then used in the cross-entropy loss for the outer optimization.

Following \cite{chen2023plot} and \cite{coop}, we choose the last training epoch model. The PLOT baseline empirically found 4 to be the optimal number of prompt features. We follow the same for our experiment. We also keep the neural network architecture and hyperparameters the same as in PLOT. For our experiment, we choose $\lambda$, kernel type and the kernel hyperparameter used in COT. We choose the featurizer in Figure~(\ref{prompt-diag}) as the same image encoder used for getting the visual features. We use a 3-layer MLP architecture for $\psi_r$ in equation \ref{cot-prompt}. We choose $\lambda$ from \{1, 10, 100\}, kernel type from {$k(x, y) = \exp{\frac{-\|x-y\|^2}{2\sigma^2}}$ (referred as RBF), $k(x, y)=(\sigma^2+\|x-y\|^2)^{-0.5}$ (referred as IMQ), $k(x, y) = \left(\frac{1+\|x-y\|^2}{\sigma^2} \right)^{-0.5}$(referred as IMQ2)}, kernel hyperparameter ($\sigma^2$) from \{median, 0.01, 0.1, 1, 10, 100\}. The chosen hyperparameters, ($\lambda$, kernel type, kernel hyperparameter), for the increasing number of shots (1 to 8), are (100, RBF, 10), (100, IMQ2, 1), (10, IMQ, 1), (1, IMQ, 0.01).

\begin{figure}[t]
    \centering
    \includegraphics[width=\columnwidth]{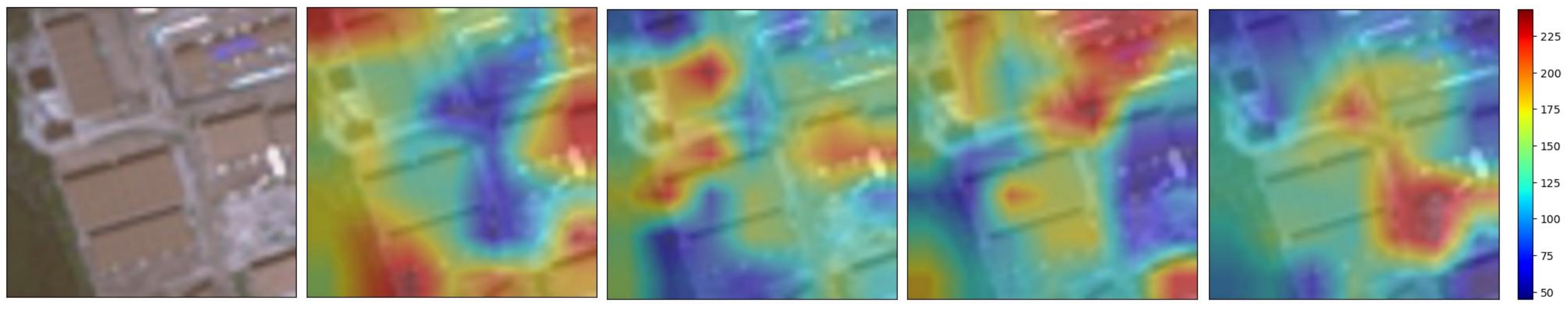}
    \caption{The leftmost is an image from the EuroSAT satellite dataset followed by visualization maps corresponding to each of the 4 prompts learnt (using COT loss \ref{cot-prompt}). We can see that the 4 prompts diversely capture different visual features of the image.}
    \label{fig:prompt-prop}
\end{figure}

\begin{figure}
    \centering
    \includegraphics[width=\columnwidth]{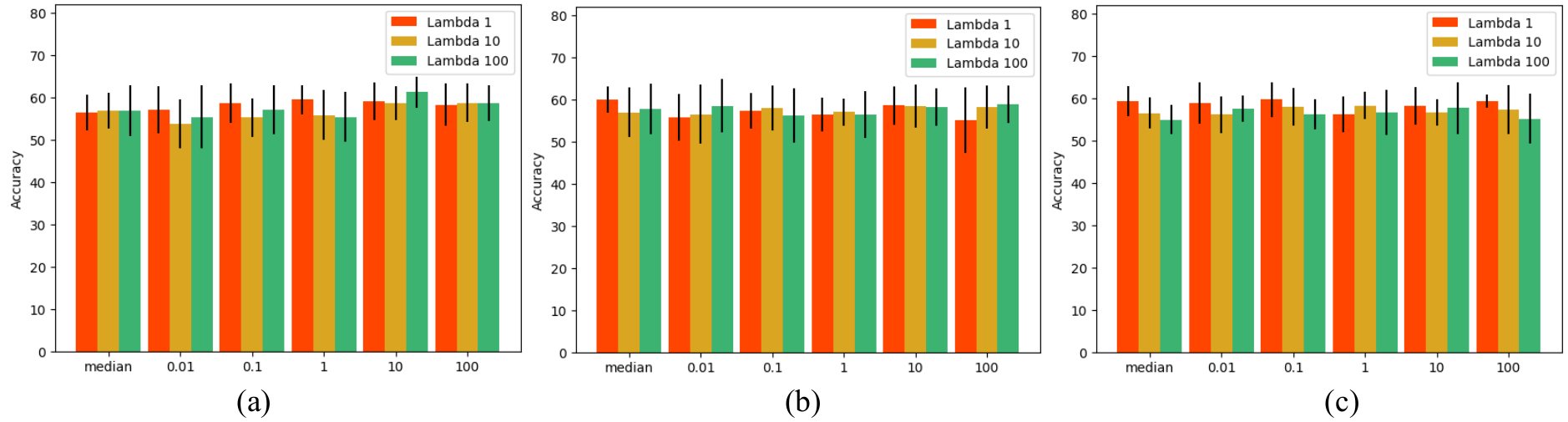}
    \caption{Ablation study for the prompt learning experiment $K=1$. For different kernel types: (a) RBF kernel $k(x, y)=\exp{\left(\frac{-\|x-y\|^2}{2\sigma^2}\right)}$ (b) IMQ kernel $k(x, y)=\left(\sigma^2+\|x-y\|^2\right)^{-0.5}$ (c) IMQ kernel $k(x, y) = \left(\frac{1+\|x-y\|^2}{\sigma^2} \right)^{-0.5}$, we show the average accuracy for different kernel hyperparameters (shown on the x-axis) and different lambda values.}
    \label{tbl:prompt-abl}
\end{figure}

Figure \ref{fig:prompt-prop} shows attention maps corresponding to each of the prompts learnt by COT. Table \ref{tbl:prompt-abl} presents an ablation study.

\section*{S3 MORE DETAILS}
\subsection*{S3.1 Motivation for the Use of MMD} As the abstract motivates, the main challenge in formulating OT over conditionals is the unavailability of the conditional distributions, which is handled by COT using MMD-based kernelized-least-squares terms computed over the joint samples that implicitly match the transport plan’s marginals with the empirical conditionals. This results in the equivalence between Eqn (4) and Eqn (5).
Furthermore, the statistical efficiency of MMD (Lemma 1) helps derive the consistency result (Thm. 1). Moreover, as discussed in $\S$ 4.2, the MMD metric is meaningful even for distributions with potentially non-overlapping support, enabling us to model the transport plan with implicit models for applications like those in $\S$ 5.2. Finally, the closed-form expression for MMD (discussed in $\S$ 2) helps in computational efficiency.
\subsection*{S3.2 The Choice of Baselines} Other baselines for $\S$ 5.1.1 and 5.1.2: CKB and CondOT are inapplicable to Fig 2. CondOT requires multiple samples for each conditioned variable ($\S$ 3 and Table 1). Using CKB, Wasserstein distance conditioned at an $x$ can't be computed, which is needed for $\S$ 5.1.1. Also, it does not provide an OT plan/map needed for $\S$ 5.1.2. Hence, these are inapplicable. We will add this clarification in $\S$ 5.
For the downstream applications in $\S$ 5.2 and $\S$ 5.3, we compare with the state-of-the-art baselines. However, for completeness's sake, we extended other baselines to these applications. The results obtained by \cite{Tabak21} for Table 2 are $\scriptstyle(7.1758, ~56.682,~559.42,~5588.14)$, for Table 3 are $\scriptstyle(0.2438,~0.587,~0.582,~0.600)$ and for Table 4 are $\scriptstyle 0.49${(MNIST)}, $\scriptstyle 0.52${(CIFAR10)}, $\scriptstyle0.52${(AWA)}. As Tables 2 and 3 need an OT map, CKB can't be applied. CondOT doesn't apply to Table 4 as they need multiple samples for each conditioned variable. Table 5 results with (\cite{Tabak21}, CKB, CondOT) are: $\scriptstyle (29.13\pm0.90, ~29.7\pm2.41,~ 23.97\pm0.98)$ for $K=1$, $\scriptstyle(38.87\pm2.00, ~26.1\pm 5.31, ~21.8\pm6.39)$ for $K=2$, $\scriptstyle(33.07\pm1.94,~28.87\pm1.58,~22.3\pm 6.98)$ for $K=4$ and $\scriptstyle(32.10\pm0.49, ~27.3\pm4.61,~21.67\pm 3.09)$ for $K=8$. The results in the manuscript ($\S$ 5) can be seen better than the above newly added.

\section*{S4 NEGATIVE SOCIETAL IMPACT}
We present a formulation for solving optimal transport between conditional distributions. This problem has many socially beneficial applications, like predicting cell responses to cancer treatment, as shown in our paper. However, if a malicious task is selected, the proposed COT formulation may have a negative societal impact, similar to most other methods in machine learning.

\end{document}